\pdfoutput=1

\documentclass[11pt]{article}

\usepackage[]{acl}

\usepackage{times}
\usepackage{latexsym}

\usepackage[T1]{fontenc}

\usepackage[utf8]{inputenc}

\usepackage{microtype}

\usepackage{inconsolata}

\usepackage{times}
\usepackage{latexsym}
\usepackage{amsmath}
\usepackage{amsfonts}
\usepackage{amssymb}
\usepackage{blindtext}
\usepackage{bbm}
\usepackage[inline]{enumitem}
\setlist{nosep}
\usepackage{pifont}
\usepackage{graphicx}
\usepackage[export]{adjustbox}
\usepackage{arydshln}
\usepackage{xcolor,colortbl}
\usepackage{adjustbox}
\usepackage{floatrow}
\usepackage{todonotes}
\usepackage{multirow}
\usepackage{soul}
\usepackage{booktabs}
\makeatletter
\newcommand\footnoteref[1]{\protected@xdef\@thefnmark{\ref{#1}}\@footnotemark}
\makeatother

\usepackage{caption}
\usepackage{subcaption}
\usepackage{bbm}
\newcommand\sbullet[1][.5]{\mathbin{\vcenter{\hbox{\scalebox{#1}{$\bullet$}}}}}
\usepackage[T1]{fontenc}

\usepackage[utf8]{inputenc}

\usepackage{microtype}

\usepackage{inconsolata}

\newcommand{\dataset}{\texttt{MCC}}
\newcommand{\proposed}{\texttt{MIME}}
\newcommand{\expmeme}{{\scshape Meme\textit{X}}}

\title{\expmeme: Detecting Explanatory Evidence for Memes\\ via Knowledge-Enriched Contextualization}

\author{Shivam Sharma$^{1,2}$, Ramaneswaran S$^{3}$, Udit Arora$^{4}$, \\\textbf{Md. Shad Akhtar$^4$ and Tanmoy Chakraborty$^1$}\\
  $^1$Indian Institute of Technology Delhi, India   \\
  $^2$Wipro AI Labs, India $\sbullet[.75]$ $^{3}$ Vellore Institute of Technology, India \\
  $^4$Indraprastha Institute of Information Technology Delhi, India \\
  \footnotesize\texttt{shivam.sharma@ee.iitd.ac.in}, \footnotesize\texttt{s.ramaneswaran2000@gmail.com}, \\\footnotesize\texttt{\{udit18417, shad.akhtar\}@iiitd.ac.in}, \footnotesize\texttt{tanchak@iitd.ac.in}
  }

\begin{document}
\maketitle
\begin{abstract}
Memes are a powerful tool for communication over social media. 
Their affinity for evolving across politics, history, and sociocultural phenomena makes them an ideal communication vehicle.
To comprehend the subtle message conveyed within a meme, one must understand the background that facilitates its holistic assimilation. Besides digital archiving of memes and their metadata by a few websites like \url{knowyourmeme.com}, currently, there is no efficient way to deduce a meme's context dynamically. In this work, we propose a novel task, \expmeme\ 
-- given a meme and a related document, the aim is to mine the context that succinctly explains the background of the meme. At first, we develop {\scshape \dataset} (Meme Context Corpus), a novel dataset for \expmeme. Further, to benchmark \dataset, we propose \proposed\ (MultImodal Meme Explainer), a multimodal neural framework that uses common sense enriched meme representation and a layered approach to capture the cross-modal semantic dependencies between the meme and the context. \proposed\ surpasses several unimodal and multimodal systems and yields an absolute improvement of $\approx 4\%$ F1-score over the best baseline. Lastly, we conduct detailed analyses of \proposed's performance, highlighting the aspects that could lead to optimal modeling of cross-modal contextual associations.
\end{abstract}

\section{Introduction}
Social media has become a mainstream communication medium for the masses, redefining how we interact within society. The information shared on social media has diverse forms, like text, audio, and visual messages, or their combinations thereof. A meme is a typical example of such social media artifact that is usually disseminated with the flair of sarcasm or humor. While memes facilitate convenient means for propagating complex social, cultural, or political ideas via visual-linguistic semiotics, they often abstract away the contextual details that would typically be necessary for the uninitiated. Such contextual knowledge is critical for human understanding and computational analysis alike.  
We aim to address this requirement by contemplating solutions that facilitate the automated derivation of contextual \textit{evidence} towards making memes more accessible.  
To this end, we formulate a novel task -- \textbf{\expmeme}, which, given a meme and a related context, aims to detect the sentences from within the context that can potentially explain the meme. Table \ref{tab:task-example} visually explains \expmeme. Memes often camouflage their intended meaning, suggesting \expmeme's utility for a broader set of multimodal applications having visual-linguistic dissociation. Other use cases include context retrieval for \textit{various art forms, news images, abstract graphics for digital media marketing}, etc.

\begin{table}[t!]
    \centering
    \resizebox{\columnwidth}{!}{%
    \setlength\tabcolsep{1.5pt} 
    \renewcommand{\arraystretch}{0.8} 
    \begin{tabular}{cm{3.5cm}}
    \hline
    
    \raisebox{-0.1cm}{\multirow{3}{*}{
    
    \includegraphics[
    width=3.4cm]{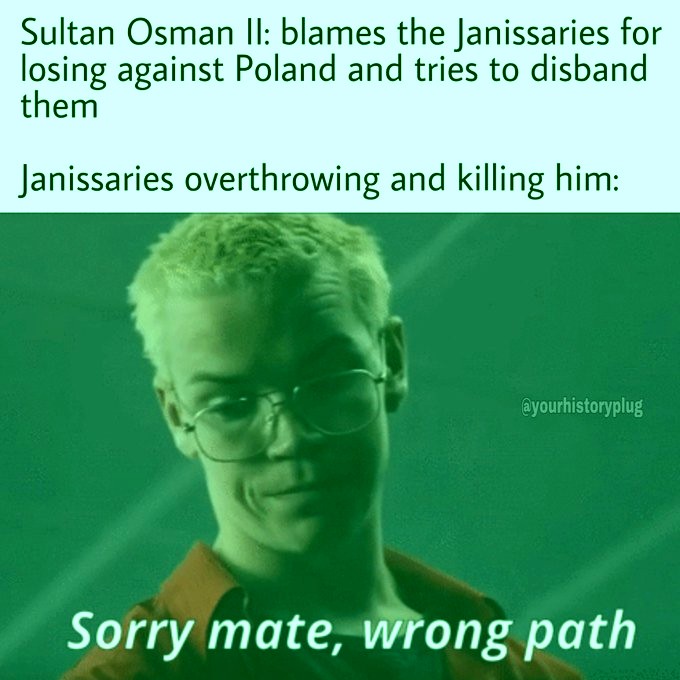}}} & \multicolumn{1}{l}{\tiny Meme Source: Reddit - \texttt{r/historymemes}}\\
    & \multicolumn{1}{l}{\tiny Context Source: Wikipedia}\\
    & \tiny \textcolor{blue}{\textbf{\hl{The result was a palace uprising by the Janissaries, who promptly imprisoned the young sultan in Yedikule Fortress in Istanbul, where Osman II was strangled to death.}}} After Osman's death, his ear was cut off and represented to Halime Sultan and Sultan Mustafa I to confirm his death and Mustafa would no longer need to fear his nephew. It was the first time in the Ottoman history that a Sultan was executed by the Janissaries.  \\ \hline
    \end{tabular}
    \caption{\expmeme\ -- given a meme and a relevant context, the aim is to identify the \textit{evidence} in the context that can succinctly explain the background of the meme, depicted above via emboldened and highlighted excerpt.}
    \label{tab:task-example}}
\end{table}

Table \ref{tab:task-example}
primarily showcases a meme's figure (left) and an excerpt from the related context (right). This meme is about the revenge killing of an \textit{Ottoman Sultan}, by the \textit{Janissaries} (infantry units), in reaction to their disbanding, by the Sultan. The first line conveys the supporting evidence for the meme from the related context, emboldened and highlighted in Table \ref{tab:task-example}. The aim is to model the required cross-modal association that facilitates the detection of such supporting pieces of evidence from a given related contextual document.


The recent surge in the dissemination of memes has led to an evolving body of studies on meme analysis in which the primary focus has been on tasks, such as emotion analysis \cite{sharma-etal-2020-semeval}, visual-semantic role labeling \cite{sharma-etal-2022-findings}, detection of phenomena like sarcasm, hate-speech \cite{kiela2020hateful}, trolling \cite{hegde2021images} and harmfulness \cite{pramanick-etal-2021-momenta-multimodal,sharma-etal-2022-disarm}. 




These studies indicate that off-the-shelf multimodal models, which perform well on several traditional visual-linguistic tasks, struggle when applied to memes \cite{kiela2020hateful,mmmlsurvey2017,sharma-etal-2022-disarm}. The primary reason behind this is the contextual dependency of memes for their accurate assimilation and analysis. Websites like \url{knowyourmeme.com} (KYM) facilitate important yet restricted information. \expmeme\ requires the model to learn the cross-modal analogies shared by the contextual evidence and the meme at various levels of information abstraction, towards detecting the crucial explanatory evidence\footnote{A comparative analysis for KYM and \proposed\ is presented in Appendix \ref{sec:appkymmime}.}. The critical challenge is to represent the abstraction granularity aptly. Therefore, we formulate \expmeme\ as an ``evidence detection'' task, which can help deduce pieces of contextual evidence that help bridge the abstraction gap. However, besides including image and text modality, there is a critical need to inject contextual signals that compensate for the constraints due to the visual-linguistic grounding offered by conventional approaches.

Even with how effective and convenient memes are to design and disseminate over social media strategically, they are often hard to understand or are easily misinterpreted by the uninitiated, typically without the proper context. Thereby suggesting the importance of addressing a task like \expmeme. Governments or organizations involved in content moderation over social media platforms could use such a utility, underlining the convenience that such a context deduction solution would bring about in assimilating harmful memes and thereby adjudicating their social implications in emergencies like elections or a pandemic.

Motivated by this, we first curate \dataset,\ a new dataset that captures various memes and related contextual documents. We also systematically experiment with various multimodal solutions to address \expmeme, which culminates into a novel framework named \proposed\ (MultImodal Meme Explainer). Our model primarily addresses the challenges posed by the knowledge gap and multimodal abstraction and delivers optimal detection of contextual evidence for a given pair of memes and related contexts. In doing so, \proposed\ surpasses several competitive and conventional baselines.     

To summarize, we make the following main contributions
\footnote{The \dataset\ dataset and the source code can be found at the URI: \url{https://github.com/LCS2-IIITD/MEMEX_Meme_Evidence.git}}.:


\begin{itemize}[leftmargin=*]
  \item {\bf A novel task}, \expmeme,\ aimed to identify explanatory evidence for memes from their related contexts.
  \item {\bf A novel dataset}, \dataset, containing $3400$ memes and related context, along with gold-standard human annotated evidence sentence-subset.     
  \item {\bf A novel method}, \proposed\ that uses common sense-enriched meme representation to identify evidence from the given context. 
  \item {\bf Empirical analysis} establishing \proposed's superiority over various unimodal and multimodal baselines, adapted for the \expmeme\ task.
\end{itemize}

\section{Related Work}
This section briefly discusses relevant studies on meme analysis that primarily attempt to capture a meme's affective aspects, such as \textit{hostility} and \textit{emotions}. Besides these, we also review other popular tasks to suitably position our work alongside different related research dimensions being explored.

\paragraph{Meme Analysis:}
Several shared tasks have been organized lately, a recent one on detecting heroes, villains, and victims from memes \cite{sharma-etal-2022-findings}, which was later followed-up via an external knowledge based approach in \cite{sharma-etal-2023-characterizing} and further extended towards generating explanations in \cite{sharma2022meme}. Other similar initiatives include troll meme classification  \cite{suryawanshi-chakravarthi-2021-findings} and meme-emotion analysis via their sentiments, types and intensity prediction \cite{sharma-etal-2020-semeval}. Notably, hateful meme detection was introduced by \citet{kiela2020hateful} and later followed-up by \citet{zhou2021multimodal}. Significant interest was garnered as a result of these, wherein various models were developed. A few efforts included fine-tuning Visual BERT \cite{li2019visualbert}, and UNITER \cite{chen2020uniter}, along with using Detectron-based representations \cite{velioglu2020detecting,lippe2020multimodal} for hateful meme detection. On the other hand, there were systematic efforts involving unified and dual-stream encoders using Transformers \cite{muennighoff2020vilio,vaswani2017attention}, ViLBERT, VLP, UNITER \cite{sandulescu2020detecting,lu2019vilbert,zhou2020unified,chen2020uniter},  and LXMERT \cite{tan2019lxmert} for dual-stream ensembling. Besides these, other tasks addressed anti-semitism \cite{chandra2021subverting}, propaganda techniques \cite{dimitrov-etal-2021-detecting}, harmfulness \cite{pramanick-etal-2021-momenta-multimodal}, and harmful targets in memes \cite{sharma-etal-2022-disarm}.

\paragraph{Visual Question Answering (VQA):}
Early prominent work on VQA with a framework encouraging \textit{open-ended} questions and candidate answers was done by \citet{VQA}. Since then, there have been multiple variations observed. \citet{VQA}  classified the answers by jointly representing images and questions. 
Others followed by examining cross-modal interactions via attention types not restricted to {co/soft/hard-attention} mechanisms \cite{VQA_Parikh1,anderson2018bottom,eccvattention2018}, effectively learning the explicit correlations between question tokens and localised image regions. Notably, there was a series of attempts toward incorporating common-sense reasoning
\cite{zellers2019recognition,wu2016ask,wu2017image,marino2019ok}. Many of these studies also leveraged information from external knowledge bases for addressing VQA tasks. 
General models like UpDn \cite{anderson2018bottom} and 
LXMERT \cite{tan2019lxmert} explicitly leverage non-linear transformations and Transformers for the VQA task, while others like LMH \cite{clark-etal-2019-dont} and SSL \cite{zhu2020prior}  addressed the critical language priors constraining the VQA performances, albeit with marginal enhancements.

\paragraph{Cross-modal Association:}
Due to an increased influx of multimodal data, the cross-modal association has recently received much attention. For cross-modal retrieval and vision-language pre-training, accurate measurement of cross-modal similarity is imperative. Traditional techniques primarily used concatenation of modalities, followed by self-attention towards learning cross-modal alignments \cite{cmr2016survey}. Following the object-centric approaches, \citet{zeng2021multi} and \citet{li2020oscar} proposed a multi-grained alignment approach, which captures the relation between visual concepts of multiple objects while simultaneously aligning them with text and additional meta-data. On the other hand, several methods also learned alignments between coarse-grained features of images and texts while disregarding object detection in their approaches \cite{huang2020pixel, kim2021vilt}. Later approaches attempted diverse methodologies, including cross-modal semantic learning from visuals and contrastive loss formulations \cite{yuan2021florence,jia2021scaling,clip2021radford}. 

Despite a comprehensive coverage of cross-modal and meme-related applications in general, 
 there are still several fine-grained aspects of memes like \textit{memetic contextualization} that are yet to be studied. 
Here, we attempt to address one such novel task, \expmeme.

\section{\dataset: Meme Context Corpus}
Due to the scarcity of publicly-available large-scale datasets that capture memes and their contextual information, we build a new dataset, \dataset\ (Meme Context Corpus). The overall dataset curation was conducted in three stages: (i) meme collection, (ii) content document curation, and (iii) dataset annotation. 
These stages are detailed in the remaining section.

\subsection{Meme Collection}
\label{sssec:meme_collection}

In this work, we primarily focus on \textit{political} and \textit{historical}, \textit{English language} memes. The reason for such a choice is the higher presence of online memes based on these topics. This is complemented by the availability of systematic and detailed information documented over well-curated digital archives. In addition to these categories, we also extend our search-space to some other themes pertaining to \textit{movies, geo-politics} and \textit{entertainment} as well. For scraping the meme images, we mainly leverage Google Images\footnote{{\scriptsize\url{https://www.google.com/imghp}}} and Reddit\footnote{\scriptsize{\url{https://www.reddit.com/}}}, for their extensive search functionality and diverse multimedia presence.

\subsection{Context Document Curation}
\label{sssec:document_collection}

We curate contextual corpus corresponding to the memes collected in the first step. This context typically constitutes pieces of evidence for the meme's background, towards which we consider Wikipedia\footnote{{\scriptsize\url{https://www.wikipedia.org/}}} (\textit{Wiki}) as a primary source. We use a Python-based wrapper API\footnote{{\scriptsize\url{https://github.com/goldsmith/Wikipedia}}} to obtain text from Wikipedia pages. For example, for \textit{Trump}, we crawl his Wiki. page \footnote{{\scriptsize\url{https://en.wikipedia.org/wiki/Donald_Trump}}}. For the scenarios wherein sufficient details are not available on a page, we look for fine-grained Wiki topics or related \textit{non-Wiki} news articles. 
For several other topics, we explore community-based discussion forums and question-answering websites like Quora\footnote{{\scriptsize\url{https://www.quora.com/}}} or other general-purpose websites.


\begin{table}[t]
\centering
\caption{Prescribed guidelines for \dataset's annotation.}
\resizebox{0.8\columnwidth}{!}{%
\begin{tabular}{cp{8cm}}
\toprule
& \multicolumn{1}{c}{\textbf{Annotation Guidelines}} \\
\midrule
1 & Meme and the associated context should be understood before annotation.\\
2 & Meme's semantics must steer the annotation.\\
3 & Self-contained, minimal units of information can constitute evidence.\\
4 & Valid evidence may or may not occur contiguously.\\   
5 & Cases not supported by a contextual document should be searched on other established sources.\\
6 & Ambiguous cases should be skipped.\\ 
\bottomrule
\end{tabular}%
}
\label{tab:annotations}
\end{table}

\begin{figure}[t!]
     \centering
     \begin{subfigure}[b]{0.49\columnwidth}
         \centering
         \includegraphics[width=\textwidth]{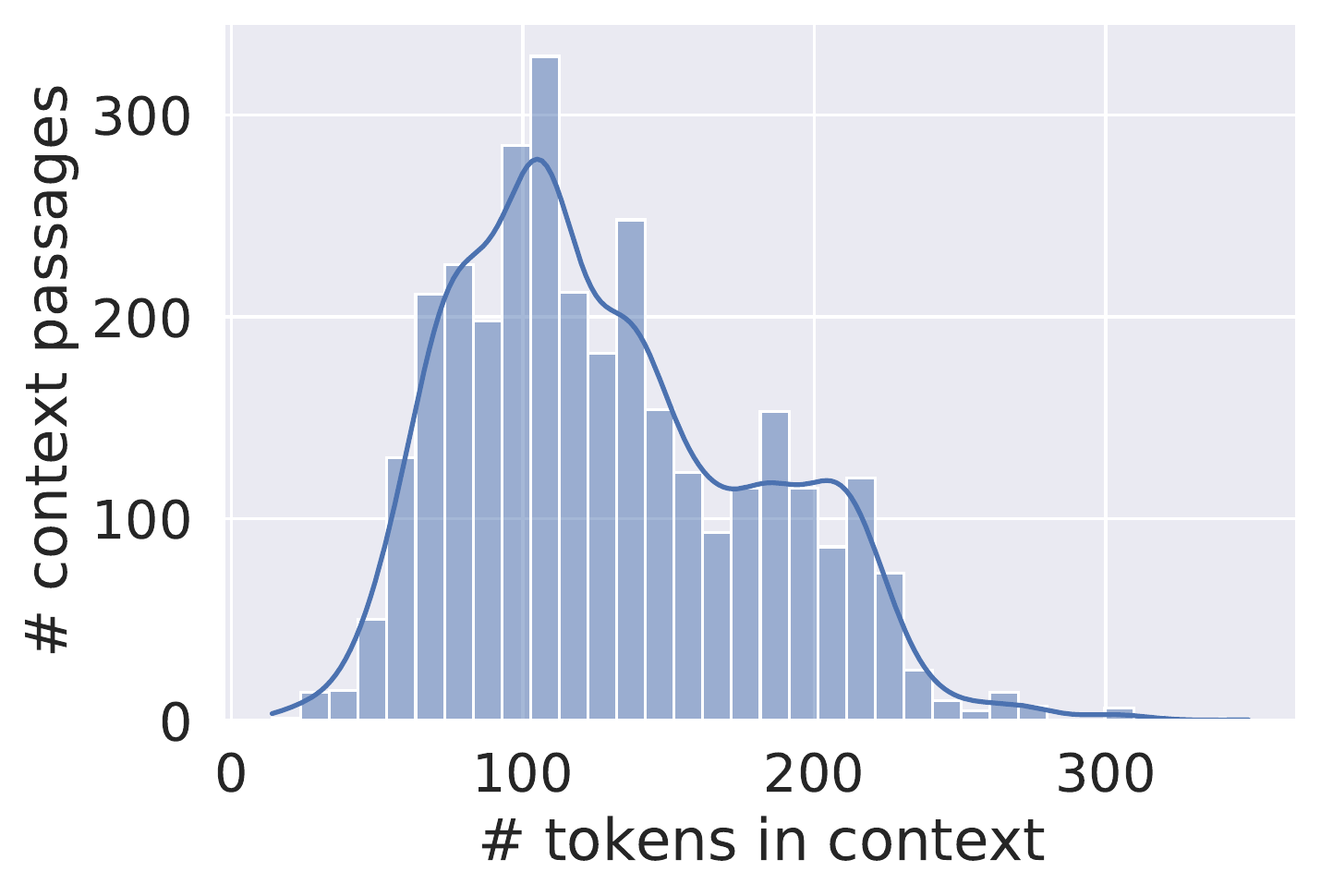}
         \caption{Context size distribution}
         \label{fig:cont-tok-dist}
     \end{subfigure}
     \hfill
     \begin{subfigure}[b]{0.49\columnwidth}
         \centering
         \includegraphics[width=\textwidth]{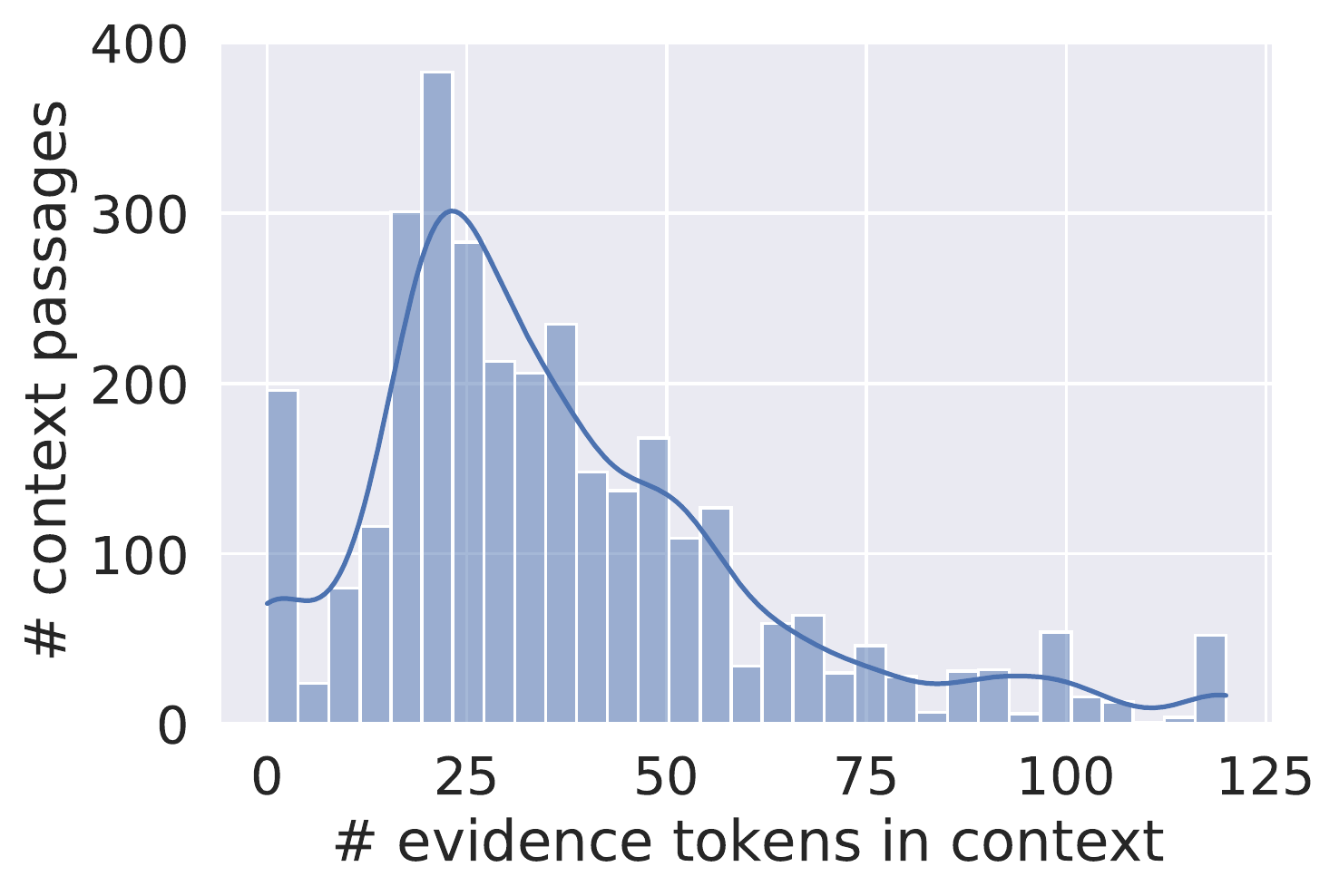}
         \caption{Evidence size distribution}
         \label{fig:ev-tok-dist}
     \end{subfigure}
        \caption{Distribution of \# tokens ($n$) in \dataset\ for: (a) related contexts ($n\in [14,349]$) and (b) context evidences ($n\in [5,312]$) (outliers $>125$, not depicted).}
        \label{fig:mcc-tok-dist}
\end{figure}
\subsection{Annotation Process}
\label{sec:annotation}


Towards curating \dataset, we employed \textit{two} annotators, one male and the other female (both Indian origin), aged between 24 to 35 yrs, who were duly paid for their services, as per Indian standards. Moreover, both were professional lexicographers and social media savvy, well versed in the urban social media vernacular. A set of prescribed guidelines for the annotation task, as shown in Table \ref{tab:annotations}, were shared with the annotators. Once the annotators were sure that they understood the meme's background, they were asked to identify the sentences in the context document that succinctly provided the background for the meme. We call these sentences ``evidence sentences'' as they facilitate (sub-)optimal \textit{evidences} that constitute likely background information. The annotation quality was assessed using \textit{Cohen's Kappa}, after an initial dry-run and the final annotation. The \textit{first} stage divulged a \textit{moderate} agreement score of 0.55, followed by several rounds of discussions, leading to a \textit{substantial} agreement score of 0.72.

\subsection{Dataset Description}



The topic-wise distribution of the memes reflects their corresponding availability on the web. Consequently, \dataset\ proportionately includes History (38.59\%), Entertainment (15.44\%), Joe Biden (12.17\%), Barack Obama (9.29\%), Coronavirus (7.80\%), Donald Trump (6.61\%), Hillary Clinton (6.33\%), US Elections (1.78\%), Elon Musk (1.05\%) and Brexit (0.95\%). Since the contextual document-size corresponding to the memes was significantly large (on average, each document consists of 250 sentences), we ensured tractability within the experimental setup by limiting the scope of the meme's related context to a subset of the entire document.
Upon analyzing the token distribution for the ground-truth pieces of evidence, we observe the maximum token length of 312 (c.f. Fig. \ref{fig:ev-tok-dist} for the evidence token distribution). Therefore, we set the maximum context length threshold to $512$ tokens. This leads to the consideration of an \textit{average} of $\approx 128$ tokens and a \textit{maximum} of over $350$ tokens (spanning $2$-$3$ paragraphs) within contextual documents (c.f. Fig. \ref{fig:cont-tok-dist} for the context token distribution). This corresponds to a maximum of $10$ sentences per contextual document.

\begin{figure*}[t!]
\centering
\includegraphics[width=0.9\textwidth]{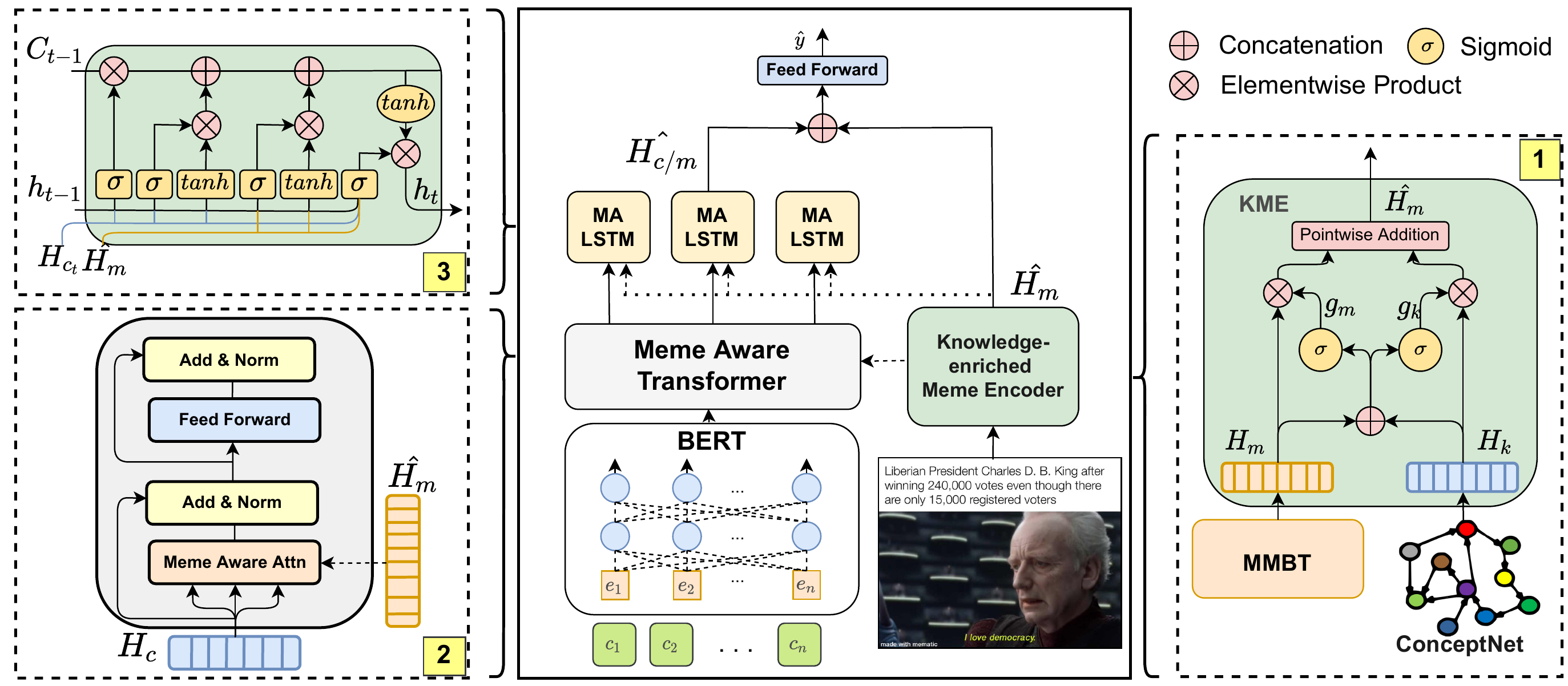}
\caption{\label{fig:main-arch}The architecture of our proposed model, \proposed. We obtain external knowledge-enriched multimodal meme representation using Knowledge-enriched Meme Encoder (KME \colorbox{yellow}{1}). We make use of a Meme-Aware Transformer (MAT \colorbox{yellow}{2}) and a Meme-Aware LSTM layer (MA-LSTM \colorbox{yellow}{3}) to incorporate the meme information while processing the context smoothly.}
\end{figure*}

We split the dataset into $80$:$10$:$10$ ratio for train/validation/test sets, resulting in $3003$ memes in the \textit{train} set and $200$ memes each in \textit{validation} and \textit{test} sets. Moreover, we ensure proportionate distributions among the train, val and test sets. Each sample in \dataset\ consists of a meme image, the context document, OCR-extracted meme's text, and a set of ground truth evidence sentences.\footnote{\label{fn1}Additional details are included in Appendix \ref{sec:app:sec:mccextra}.} 

\section{Methodology}

In this section, we describe our proposed model, \proposed. It takes a meme (an image with overlaid text) and a related context as inputs and outputs a sequence of labels indicating whether the context's constituting \textit{evidence sentences}, either in part or collectively, explain the given meme or not. 

As depicted in Fig. \ref{fig:main-arch}, \proposed\ consists of a text encoder to encode the context and a multimodal encoder to encode the meme (image and text). 
To address the complex abstraction requirements, we design a Knowledge-enriched Meme Encoder (KME) that augments the joint multimodal representation of the meme with external common-sense knowledge via a gating mechanism. On the other hand, we use a pre-trained BERT model to encode the sentences from the candidate context. 

We then set up a Meme-Aware Transformer (MAT) to integrate meme-based information into the context representation for designing a multi-layered contextual-enrichment pipeline. Next, we design a Meme-Aware LSTM (MA-LSTM) that sequentially processes the context representations conditioned upon the meme-based representation. Lastly, we concatenate the last hidden context representations from MA-LSTM and the meme representation and use this jointly-contextualized meme representation for evidence detection. Below, we describe each component of \proposed\ in detail. 

\paragraph{Context Representation:}

Given a related context, $C$ consisting of sentences $[c_1, c_2 ... c_n]$, we encode each sentence in $C$ \textit{individually} using a pre-trained BERT encoder, and the pooled output corresponding to the \texttt{[CLS]} token is used as the context representation. Finally, we concatenate the individual sentence representation to get a unified context representation $H_{c}\in \mathbb{R}^{n\times d}$, with a total of $n$ sentences.





\paragraph{Knowledge-enriched Meme Encoder:}

Since memes encapsulate the complex interplay of linguistic elements in a contextualized setting, it is necessary to facilitate a primary understanding of linguistic abstraction besides factual knowledge. In our scenario, the required contextual mapping is implicitly facilitated across the contents of the meme and context documents. Therefore, to supplement the feature integration with the required common sense knowledge, we employ ConceptNet \cite{concept_net}: a semantic network designed to help machines comprehend the meanings and semantic relations of the words and specific facts people use. Using a pre-trained GCN, trained using ConceptNet, we aim to incorporate the semantic characteristics by extracting the averaged GCN-computed representations corresponding to the meme’s text. In this way, the representations obtained are common sense-enriched and are further integrated with the rest of the proposed solution.


To incorporate external knowledge, we use ConceptNet \cite{concept_net} knowledge graph (KG) as a source of external commonsense knowledge. 
To take full advantage of the KG and at the same time to avoid the query computation cost, we use the last layer from a pre-trained graph convolutional network (GCN), trained over ConceptNet \cite{malaviya_kg_completion}.

We first encode meme $M$ by passing the meme image $M_i$ and the meme text $M_t$\footnote{Extracted using Google Vision's OCR API: \scriptsize\url{https://cloud.google.com/vision/docs/ocr}} to an empirically designated pre-trained MMBT model \cite{mmbt}, to obtain a multimodal representation of the meme $H_m\in \mathbb{R}^{d}$.
Next, to get the external knowledge representation, we obtain the GCN node representation corresponding to the words in the meme text $M_t$. This is followed by average-pooling these embeddings to obtain the unified knowledge representation $H_k\in \mathbb{R}^{d}$.

To learn a knowledge-enriched meme representation $\hat{H}_{m}$, we design a Gated Multimodal Fusion (GMF) block. As part of this, we employ a \textit{meme gate} ($g_m$) and the \textit{knowledge gate} ($g_k$) to modulate and fuse the corresponding representations.

\begin{equation}
\small
    \begin{array}{rcl}
        g_m = \sigma([H_m + H_k]W_{m} + b_{m})\\
        g_k = \sigma([H_m + H_k]W_{k} + b_{k})
    \end{array}
\end{equation}
Here, $W_{m}$ and $W_{k} \in 	\mathbb{R}^{2d\times d}$ 
are trainable parameters. 



\paragraph{Meme-Aware Transformer:}

A conventional Transformer encoder \cite{attention_is_all_you_need} uses self-attention, which facilitates the learning of the inter-token contextual semantics. However, it does not consider any additional contextual information helpful in generating the query, key, and value representations. Inspired by the context-aware self-attention proposed by \citet{context_aware_attention}, in which the authors proposed several ways to incorporate \textit{global}, \textit{deep}, and \textit{deep-global} contexts while computing self-attention over \textit{embedded textual tokens}, we propose a meme-aware multi-headed attention (MHA). This facilitates the integration of \textit{multimodal meme information} while computing the self-attention over context representations. We call the resulting encoder a meme-aware Transformer (MAT) encoder, which is aimed at computing the cross-modal affinity for $H_c$, conditioned upon the knowledge-enriched meme representation $\hat{H}_{m}$. 



Conventional self-attention uses query, key, and value vectors from the same modality. In contrast, as part of meme-aware MHA, we first generate the key and the value vectors conditioned upon the meme information and then use these vectors via conventional multi-headed attention-based aggregation. We elaborate on the process below.


Given the context representation $H_c$, we first calculate the conventional query, key, and value vectors $Q$, $K$, $V \in \mathbb{R}^{n\times d}$, respectively as given below:
\begin{equation}
\small
    [QKV] = H_c[W_{Q}W_{K}W_{V}]
\end{equation}
Here, $n$ is the maximum sequence length, $d$ is the embedding dimension, and $W_{Q},W_{K},$ and $W_{V} \in \mathbb{R}^{d\times d}$ are learnable parameters. 

We then generate new key and value vectors $\hat{K}$ and $\hat{V}$, respectively, which are conditioned on the meme representation $\hat{H}_{m} \in \mathbb{R}^{1\times d}$ (broadcasted corresponding to the context size). 
We use a gating parameter $\lambda \in \mathbb{R}^{n\times 1}$ to regulate the memetic and contextual interaction. Here, $U_k$ and $U_v$ constitute learnable parameters.  
\begin{equation}
\small
   {\begin{bmatrix}
\hat{K}
\\  
\hat{V}
\end{bmatrix}} = (1-{\begin{bmatrix}
\lambda_k
\\  
\lambda_v
\end{bmatrix}}){\begin{bmatrix}
K
\\  
V
\end{bmatrix}} 
+ 
{\begin{bmatrix}
\lambda_k
\\  
\lambda_v
\end{bmatrix}}
(\hat{H_{m}} 
{\begin{bmatrix}
U_k
\\  
U_v
\end{bmatrix}} 
)
\end{equation}


We learn the parameters $\lambda_k$ and $\lambda_v$ using a sigmoid based gating mechanism instead of treating them as hyperparameters as follows:
\begin{equation}
\small
    {\begin{bmatrix}
\lambda_{k}
\\  
\lambda_{v}
\end{bmatrix}} = \sigma({\begin{bmatrix}
K
\\  
V
\end{bmatrix}} 
{\begin{bmatrix}
W_{k_1}
\\  
W_{v_1}
\end{bmatrix}} 
+
\hat{H_{m}}  
{\begin{bmatrix}
U_k
\\  
U_v
\end{bmatrix}}
{\begin{bmatrix}
W_{k_2}
\\  
W_{v_2}
\end{bmatrix}}
)
\end{equation}
Here, $W_{k_1}$, $W_{v_1}$, $W_{k_2}$ and $W_{v_2} \in \mathbb{R}^{d\times 1}$ are learnable parameters.



Finally, we use the query vector $Q$ against $\hat{K}$ and $\hat{V}$, conditioned on the meme information in a conventional scaled dot-product-based attention. This is extrapolated via multi-headed attention to materialize the Meme-Aware Transformer (MAT) encoder, which yields meme-aware context representations $H_{c/m}\in \mathbb{R}^{n\times d}$. 


\paragraph{Meme-Aware LSTM:}

Prior studies have indicated that including a recurrent neural network such as an LSTM with a Transformer encoder like BERT is advantageous. Rather than directly using a standard LSTM in \proposed, we aim to incorporate the meme information into sequential recurrence-based learning. Towards this objective, we introduce Meme-Aware LSTM (MA-LSTM) in  \proposed. MA-LSTM is a recurrent neural network inspired by \cite{xu-etal-2021-better} that can incorporate the meme representation $\hat{H_{m}}$ while computing cells and hidden states. The gating mechanism in MA-LSTM allows it to assess how much information it needs to consider from the hidden states of the enriched context and meme representations, $H_{c/m}$ and $\hat{H_{m}}$, respectively. 

Fig. \ref{fig:main-arch} shows the architecture of MA-LSTM. We elaborate on the working of the MA-LSTM cell below. It takes as input the previous cell states $c_{t-1}$, previous hidden representation $h_{t-1}$, current cell input $H_{c_{t}}$, and an additional meme representation $\hat{H_m}$. Besides the conventional steps involved for the computation of \textit{input, forget, output} and \textit{gate} values w.r.t the input $H_{c_{t}}$, the \textit{input} and the \textit{gate} values are also computed w.r.t the additional input $\hat{H_{m}}$. The final \textit{cell} state and the \textit{hidden} state outputs are obtained as follows:
\begin{equation*}
\small
    \begin{array}{rcl}
    c_{t} &=& f_{t}\odot c_{t-1} + i_{t}\odot \hat{c_{t}} + p_{t}\odot \hat{s_{t}} \\
    h_{t} &=& o_t \odot tanh(c_t)
    \end{array}
\end{equation*}

The hidden states from each time step are then concatenated to produce the unified context representation $\hat{ H_{c/m}}\in \mathbb{R}^{n\times d}$.

\paragraph{Prediction and Training Objective:}

Finally, we concatenate $\hat{ H_{m}}$ and $\hat{ H_{c/m}}$ to obtain a joint context-meme representation, which we then pass through a feed-forward layer to obtain the final classification. The model outputs the \textit{likelihood} of a sentence being valid evidence for a given meme. We use the cross-entropy loss to optimize our model.

\section{Baseline Models}

We experiment with various unimodal and multimodal encoders for systematically encoding memes and context representations to establish comparative baselines. The details are presented below. 

\paragraph{Unimodal Baselines:}
\begin{itemize*}
    \item \textbf{BERT \citep{devlin2019BERT}:} To obtain text-based unimodal meme representation.
    \item \textbf{ViT \citep{vit}:} Pre-trained on ImageNet to obtain image-based unimodal meme representation. 
\end{itemize*}
\paragraph{Multimodal Baselines:}
\begin{itemize*}
    \item \textbf{Early-fusion:} To obtain a concatenated multimodal meme representation, using BERT and ViT model.
    \item \textbf{MMBT \citep{mmbt}:} For leveraging projections of pre-trained image features to text tokens to encode via multimodal bi-transformer. 
    \item \textbf{CLIP \citep{clip2021radford}:} To obtain multimodal representations from memes using CLIP image and text encoders, whereas CLIP text encoder for context representation.
    \item \textbf{BAN \citep{BAN}:}  To obtain a joint representation using low-rank bilinear pooling while leveraging the dependencies among two groups of input channels. 
    \item \textbf{VisualBERT \citep{li2019visualbert}:} To obtain multimodal pooled representations for memes, using a Transformer-based visual-linguistic model. 
\end{itemize*}

\section{Experimental Results}
\label{sec:experiments-and-results}

This section presents the results (averaged over five independent runs) on our thematically diversified \textit{test-set} and performs a comparison, followed by qualitative and error analysis. For comparison, we use the following standard metrics -- accuracy (Acc.), macro averaged F1, precision (Prec.), recall (Rec.), and exact match (E-M) score\footnote{Additional experimental details are available in Appendix \ref{sec:app:imphyper}.}. To compute the scores corresponding to the partial match scenarios, we compute the precision/recall/F1 separately for each case before averaging across the test set. Additionally, as observed in \cite{BESKOW2020102170}, we perform some basic image-editing operations like adjusting \textit{contrast, tint, temperature, shadowing} and \textit{highlight}, on meme images in \dataset\ for (i) optimal OCR extraction of meme text, and (ii) noise-resistant feature learning from images\footnote{See Section \ref{sec:ethics} for details on \textit{Terms and Conditions for Data Usage}.}.
\begin{table}[t!]
\begin{adjustbox}{width=\columnwidth,center}
\begin{tabular}{ccccccc}
\toprule
\textbf{Type} & \multicolumn{1}{c}{\textbf{Model}}    & \multicolumn{1}{c}{\textbf{Acc.}} & \multicolumn{1}{c}{\textbf{F1}} & \multicolumn{1}{c}{\textbf{Prec.}} & \multicolumn{1}{c}{\textbf{Rec.}} & \multicolumn{1}{c}{\textbf{E-M}}\\ \midrule
\multirow{2}{*}{UM} & Bert              & 0.638                                 & 0.764                                 & 0.768                                  & 0.798 & 0.485                               \\ 
 
& ViT               & 0.587                                  & 0.698                                 & 0.711                                  & 0.720                & 0.450                \\ \midrule
\multirow{6}{*}{MM} & E-F      & 0.646                                 & 0.772                                 & 0.787                                  & 0.798       & 0.495                         \\
& CLIP              & 0.592                                 & 0.709                                 & 0.732                                  & 0.747     & 0.460                           \\
& BAN & 0.638 & 0.752 & 0.767 & 0.772 & 0.475 \\ 
& V-BERT        & 0.641                                 & 0.765                                 & 0.773                                  & 0.783              & 0.490                  \\ 
& MMBT  $\dagger$            & 0.650                                 & 0.772                                 & 0.790                                  & 0.805              & 0.505                   \\ \cdashline{2-7}
& \proposed\ $^{\star}$ & \textbf{0.703}                                 & \textbf{0.812}                                 & \textbf{0.833}                                  & \textbf{0.828}     & \textbf{0.585}                           \\ \hline
\multicolumn{2}{c}{$\Delta_{\text{(\proposed\ -MMBT)}}$}             & {\textcolor{blue}{$\uparrow5.34\%$}}                                & {\textcolor{blue}{$\uparrow3.97\%$}}                                 & {\textcolor{blue}{$\uparrow4.26\%$}}                                 & {\textcolor{blue}{$\uparrow2.31\%$}}
                            & {\textcolor{blue}{$\uparrow8.00\%$}}\\ \bottomrule
\end{tabular}
\end{adjustbox}
\caption{\label{tab:main-results-table}Comparison of different approaches on \dataset. The last row shows the absolute improvement of \proposed\ over MMBT (the best baseline). E-F: Early Fusion and V-BERT: VisualBERT.}
\end{table}

\paragraph{Meme-evidence Detection (\expmeme):}As part of performance analysis, we observe from Table  \ref{tab:main-results-table} that unimodal systems, in general, perform with mediocrity, with the Bert-based model yielding a relatively better F1 score of 0.7641, as compared to the worst score of 0.6985 by ViT-based model. It can be reasoned that textual cues would be significantly pivotal in modeling association when the target modality is also text-based. On the contrary, purely image-based conditioning would not be sufficient for deriving fine-grained correlations for accurately detecting correct evidence. Also, the lower precision, as against the higher recall scores, suggests the inherent noise being additionally modeled.         

\begin{table}[t!]
    \centering
    \resizebox{\columnwidth}{!}{%
    \setlength\tabcolsep{1.5pt} 
    \renewcommand{\arraystretch}{0.6} 
    \begin{tabular}{cm{4.9cm}}
    \toprule
    
    \raisebox{-0.1cm}{\multirow{4}{*}{
    
    \includegraphics[
    height=3.9cm,
    width=2.5cm]{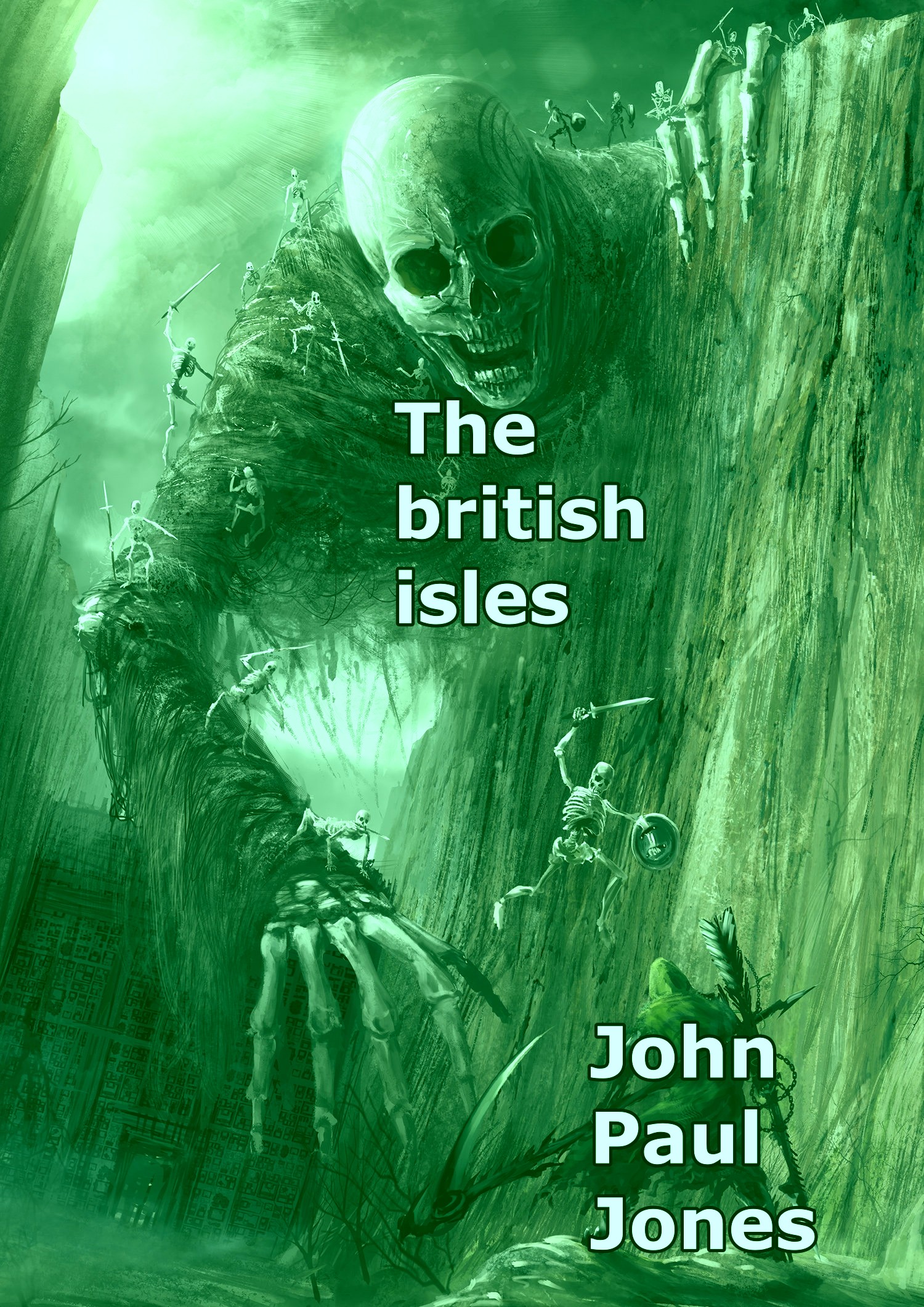}}} & \multicolumn{1}{c}{\tiny \textbf{MMBT}}\\
    & \tiny \hl{John Paul Jones was a Scottish-American naval captain who was the United States' first well-known naval commander in the American Revolutionary War.} He made many friends among U.S political elites, as well as enemies (who accused him of piracy). \textcolor{blue}{\textbf{His actions in British waters during the Revolution earned him an international reputation which persists to this day.}} \\ \cdashline{2-2}
     & \multicolumn{1}{c}{\tiny \textbf{\proposed}}\\
     & \tiny John Paul Jones was a Scottish-American naval captain who was the United States' first well-known naval commander in the American Revolutionary War. He made many friends among U.S political elites, as well as enemies (who accused him of piracy). \textcolor{blue}{\textbf{\hl{His actions in British waters during the Revolution earned him an international reputation which persists to this day.}}} \\ \bottomrule
    \end{tabular}
    \caption{Evidence detection from MMBT (top) and \proposed\ (bottom) for a sample meme. The emboldened sentences in blue indicate \textcolor{blue}{\textbf{ground-truth evidences}} and the highlighted sentences indicate \hl{model prediction}.}
    \label{tab:qual-analysis}}
\end{table}
On the other hand, multimodal models either strongly compete or outperform unimodal ones, with CLIP being an exception. With an impressive F1 score of 0.7725, MMBT fares optimally compared to the other comparative multimodal baselines. This is followed by the early-fusion-based approach and VisualBERT, with 0.7721 and 0.7658 F1 scores, respectively. BAN (Bilinear Attention Network) performs better than early-fusion and CLIP, but falls short by a 1-2\% F1 score. Models like MMBT and VisualBERT leverage pre-trained unimodal encoders like BERT and ResNet and project a systematic joint-modeling scheme for multiple modalities. Although this has proven to be beneficial towards addressing tasks that leverage visual-linguistic grounding, especially when pre-trained using large-scaled datasets like MSCOCO (VisualBERT),  their limitations can be ascertained from Table \ref{tab:main-results-table}, wherein \proposed\ yields absolute improvements of 5.34\%, 3.97\%, 4.26\%, 2.31\% and 8.00\% in accuracy, F1 score, precision, recall, and exact match scores, respectively, over the best baseline, MMBT. This suggests potential improvement that a systematic and optimal contextualization-based approach like \proposed\ can offer.

\paragraph{Analysing Detected Evidences:}

We analyze the detected evidence by contrasting \proposed's prediction quality with MMBT's. 
The meme depicted in Table \ref{tab:qual-analysis}
does not explicitly convey much information and only mentions two entities, ``John Paul Jones'' and ``The British Isles''. The MMBT baseline predicts the first sentence as an explanation, which contains the word ``John Paul Jones'', whereas \proposed\ correctly predicts the last sentence that explains the meme. Observing the plausible multimodal analogy that might have led \proposed\ to detect the relevant evidence in this case correctly is interesting. In general, 
we observe that the evidence predicted by MMBT does not fully explain the meme, whereas those predicted by \proposed\ are often more fitting.

\begin{table}[t!]
\begin{adjustbox}{width=\columnwidth,center}
\begin{tabular}{clccccc}
\toprule
\multicolumn{1}{c}{\textbf{System}} & 
\multicolumn{1}{c}{\textbf{Model}}   & \multicolumn{1}{c}{\textbf{Acc.}} & \multicolumn{1}{c}{\textbf{F1}} & \multicolumn{1}{c}{\textbf{Prec.}} & \multicolumn{1}{c}{\textbf{Rec.}} & \multicolumn{1}{c}{\textbf{E-M}} \\\midrule
\multirow{4}{*}{\rotatebox{90}{\begin{tabular}[c]{@{}c@{}}MMBT \\ \& variants\end{tabular}}} & MMBT             & 0.650                                 & 0.772                                 & 0.790                                  & 0.805                  & 0.505              \\ \cdashline{2-7}
&
 \textbf{+} KME              & 0.679                                 & 0.789                                 & 0.804                                   & 0.822               & 0.550                \\ 
 & \textbf{+} MAT              & 0.672                                 & 0.793                                 & 0.810                                  & 0.814                    & 0.540          \\ 
 & \textbf{+} MA-L          & 0.639                                 & 0.780                                 & 0.791                                  & 0.808              & 0.490                 \\ \midrule
 \multirow{5}{*}{\rotatebox{90}{\begin{tabular}[c]{@{}c@{}}\proposed\ \\ \& variants\end{tabular}}} & \textbf{--} MA-L   & 0.694                                 & 0.800                                 & 0.826                                  & 0.8234          & 0.560                     \\ 
 & \textbf{--} MA-L \textbf{+} BiL      & 0.689                                 & 0.807                                 & 0.8141                                  & 0.826 & 0.565\\ 
 & \textbf{--} MAT       & 0.649                                 & 0.783                                 & 0.788                                  & 0.811                    & 0.510           \\ 
 & \textbf{--} MAT \textbf{+} T & 0.687                                 & 0.779                                 & 0.801                                  & 0.813   & 0.560
                                 \\ \cdashline{2-7}

& \proposed\         & \textbf{0.703}                                & \textbf{0.812}                                 & \textbf{0.833}                                  & \textbf{0.828}        & \textbf{0.585}                       \\ \bottomrule
\end{tabular}
\end{adjustbox}
\caption{\label{tab:ablation-table}Component-wise evaluation: each component contributes to the performance of \proposed, while removing them inhibits it. T: Transformer, L: LSTM, BiL: Bi-LSTM and MA: Meme-Aware.}
\end{table}
\paragraph{Ablation Study:}
\label{ssec:ablation-study}

\proposed's key modules are Knowledge-enriched Meme Encoder (KME), Meme-Aware Transformer (MAT) encoder, and Meme-Aware LSTM (MA-LSTM). The incremental assessment of these components, over MMBT as a base model, can be observed from Table \ref{tab:ablation-table}. Adding external knowledge-based cues along with the MMBT representation via KME leads to an enhancement of 0.98\%-2.91\% and 5\% across the first four metrics and the exact match, respectively. Similar enhancements are observed with MAT and MA-LSTM, with increments of 0.91-2.25\% and 0.06-2.25\%, respectively. Therefore, it can be reasonably inferred that KME, MAT, and MA-LSTM distinctly contribute towards establishing the efficacy of \proposed. 

On removing MA-LSTM, we notice a distinct performance drop $\in[0.47,2.50]$\% across all five metrics. Dropping MAT from \proposed\ downgrades the performance by 1.67-5.38\% for the first four metrics and by 7.5\% for the exact match score.

Finally, we examine the influence via replacement by employing a standard Transformer-based encoder instead of MAT and a BiLSTM layer instead of MA-LSTM, in \proposed. The former results in a drop of 1.45-3.28\% across all five metrics. Whereas, the drop for the latter is observed to be 0.21\%-2.00\%. This suggests the utility of systematic memetic contextualization while addressing \expmeme.

\begin{table}[t!]
    \centering
    \resizebox{\columnwidth}{!}{%
    \setlength\tabcolsep{1.5pt} 
    \begin{tabular}{m{2cm}m{5.1cm}}
    \toprule
    \multicolumn{1}{c}{\small Meme} & \multicolumn{1}{c}{\small Related Context} \\ \midrule
    \raisebox{-0.1\height}{\includegraphics[
    width=2cm]{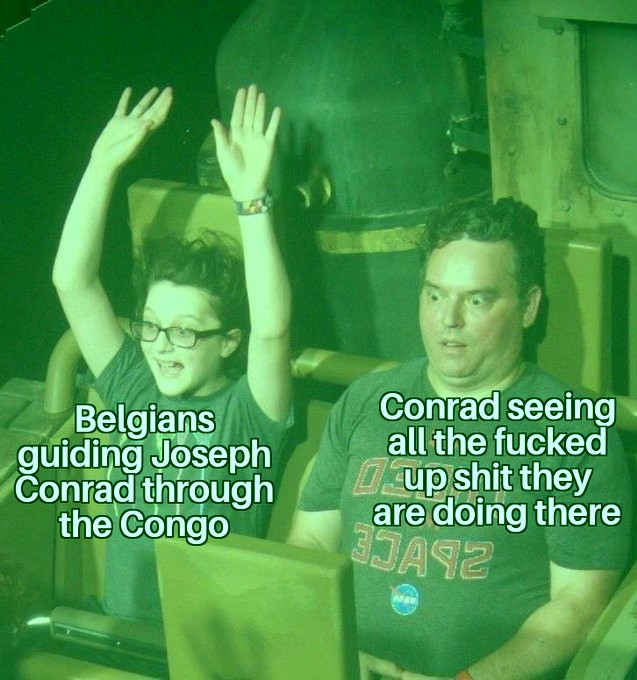}} & \tiny Heart of Darkness (1899) is a novella by Polish-English novelist Joseph Conrad. It tells the story of Charles Marlow, a sailor who takes on an assignment from a Belgian trading company as a ferry-boat captain in the African interior. \textbf{\textcolor{blue}{The novel is widely regarded as a critique of European colonial rule in Africa, whilst also examining the themes of power dynamics and morality. Although Conrad does not name the river where the narrative takes place, at the time of writing the Congo Free State, the location of the large and economically important Congo River, was a private colony of Belgium's King Leopold II.}} \\ \cdashline{1-2}
    \raisebox{-0.1\height}{\includegraphics[
    width=2cm]{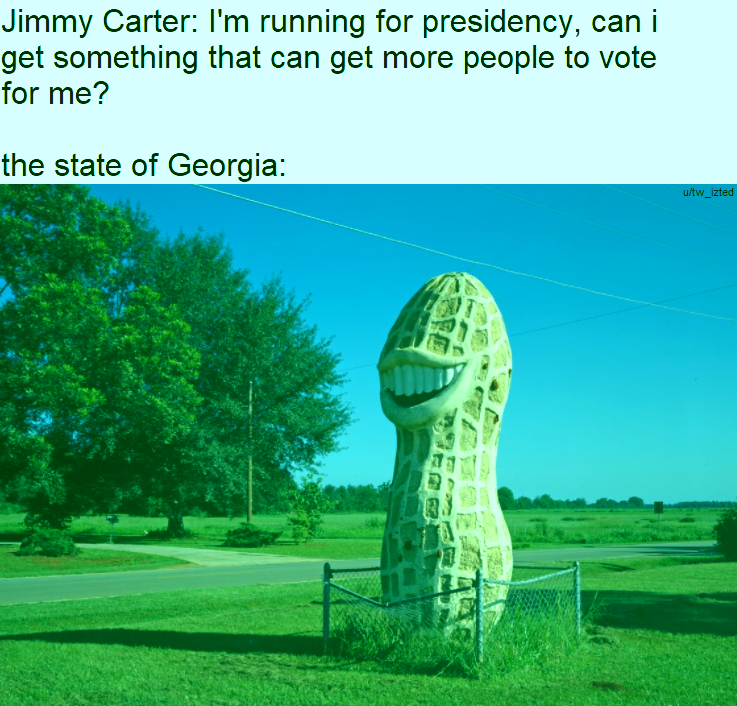}} & \tiny The Jimmy Carter Peanut Statue is a monument located in Plains, Georgia, United States. \textcolor{blue}{\textbf{\hl{Built in 1976, the roadside attraction depicts a large peanut with a toothy grin, and was built to support Jimmy Carter during the 1976 United States presidential election.}}}\hl{ The statue was commissioned by the Indiana Democratic Party during the 1976 United States presidential election as a form of support for Democratic candidate Jimmy Carter's campaign through that state. The statue, a 13-foot (4.0 m) peanut, references Carter's previous career as a peanut farmer.} \\ \cdashline{1-2}
    \raisebox{-0.1\height}{\includegraphics[
    width=2cm]{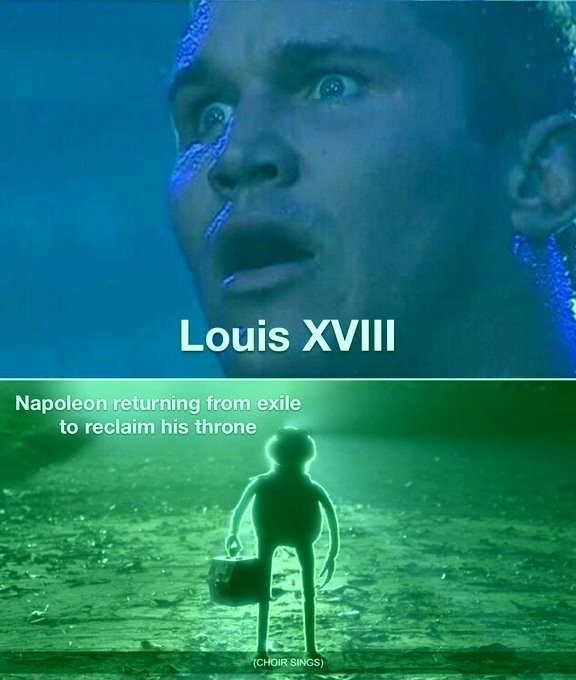}} & \tiny \hl{On February 26, 1815, Napoleon managed to sneak past his guards and somehow escape from Elba, slip past interception by a British ship, and return to France.} Immediately, people and troops began to rally to the returned Emperor. French police forces were sent to arrest him, but upon arriving in his presence, they kneeled before him. Triumphantly, Napoleon returned to Paris on March 20, 1815. \textcolor{blue}{\textbf{Paris welcomed him with celebration, and Louis XVIII, the new king, fled to Belgium. With Louis only just gone, Napoleon moved back into the Tuileries.}} The period known as the Hundred Days had begun.\\\bottomrule
    \end{tabular}
    \caption{Prediction errors from \proposed\ on three \textit{test-set} samples. The emboldened sentences in blue indicate \textcolor{blue}{\textbf{ground-truth evidences}} and the highlighted sentences indicate \hl{model prediction}.}
    \label{tab:error-examples}}
\end{table}

\paragraph{Error Analysis:}

Here, we analyze different types of errors incurred by the model. As observed from the first example in Table \ref{tab:error-examples}, ground-truth evidence contain abstract concepts like \textit{power dynamics and morality}, along with various novel facts, which induce non-triviality.
On the other hand, the second example depicts a partial prediction, wherein the extra excerpt detected by the \proposed\ is likely due to the inductive biases based on concepts of \textit{presidential race, Jimmy Carter and visual description of the peanut statue}. Finally, the model just mapped its prediction based on the embedded meme text, e.g., \#3, while partly oblivious to the meme's visuals. Overall, \proposed\ obtains an exact match for 58.50\% of the test-set cases. At the same time, it cannot predict any explanation for 12.5\% cases. The model obtains partial matches for about 14\% of the cases, and for the remaining 14\%, the model makes wrong predictions.\footnote{Further discussion is available in Appendix \ref{app:sec:limit}.}


\paragraph{Discussion:}As part of this study, we examine \proposed's efficacy over other variants when the constituting components are considered both incrementally and decrementally (c.f Table \ref{tab:ablation-table}). Notably, we observe that adding external common sense knowledge-based signals, and attending over the meme while processing the context evidence sentences using MAT and MA-LSTM modules, distinctly increases the performance. These components are empirically observed and demonstrated to induce performance enhancement and establish their efficacy proving their respective hypotheses of augmenting the representation learning with common sense-based multimodal feature enrichment, self-attention-based multimodal Transformer encoding of the pieces of evidence, and finally, sequence modeling of the derived multimodal Transformer representations, modeling their temporal entailment embedded in their contextual arrangement.

To further delineate the scope of this study, it does not aim to deduce/derive every possible contextual evidence that can comprehensively contextualize a given meme; instead, it is to derive the evidence pieces, given closely related raw information (which can be conveniently obtained by directed query searches), that can help provide that necessary contextual impetus towards adjudicating various memetic phenomenon (like hate, offense, etc.). The fact that such a pipeline is not constrained by a particular topic, domain, and information source makes it reasonably scalable.

\section{Conclusion}

This work proposed a new task -- \expmeme\ that aims to identify evidence from a given context to explain the meme. To support this task, we also curated \dataset,\ a novel manually-annotated multimodal dataset encompassing a broad range of topics. After that, we benchmarked \dataset\ on several competitive systems and proposed \proposed,\ a novel modeling framework that utilizes knowledge-enriched meme representation and integrates it with context via a unique multi-layered fusion mechanism. The empirical examination and an extensive ablation study suggested the efficacy of \proposed\ and its constituents. We then analyzed \proposed's correct contextual mapping heuristics, juxtaposed with its limitations, suggesting the possible scope of improvement. 

\section*{Limitations}

\label{app:sec:limit}
Although our approach, \proposed\ is empirically observed to outperform several other competitive baselines, we do observe some limitations in the modeling capacity towards \expmeme. As depicted in Table \ref{tab:error-examples}, there are three possible scenarios of ineffective detection -- (a) no predictions, (b) partial match, and (c) incorrect predictions. The key challenges stem from the limitations in modeling the complex level of abstractions that a meme exhibits. These are primarily encountered in either of the following potential scenarios:
\begin{itemize}[leftmargin=*]
    \item A critical, yet a cryptic piece of information within memes, comes from the visuals, which typically requires some systematic integration of factual knowledge, that currently lacks in \proposed.
    \item Insufficient textual cues pose challenges for \proposed,\ for learning the required contextual associativity.
    \item Potentially spurious pieces of evidence being picked up due to the lexical biasing within the related context.
\end{itemize}


\section*{Ethics and Broader Impact}
\label{sec:ethics}

\paragraph{Reproducibility.}  We present detailed hyper-parameter configurations in Appendix~\ref{sec:app:imphyper} and Table~\ref{tab:hyper-table}. 
The source code and \dataset\ dataset are publicly shared at \url{https://github.com/LCS2-IIITD/MEMEX_Meme_Evidence.git}. 
 
\paragraph{Data Collection.} 
The data collection protocol was duly approved by an ethics review board. 
 
\paragraph{User Privacy.} 
The information depicted/used does not include any personal information. 


\paragraph{Terms and Conditions for data usage.}
We  performed basic image editing (c.f. Section \ref{sec:experiments-and-results}) on the meme images downloaded from the Internet and used for our research. This ensures non-usage of the artwork/content in its original form. 

Moreover, we  already included details of the subreddits and keywords used to collect meme content and the sources used for obtaining contextual document information as part of Appendix \ref{subsec:MemeCollectApp}, Section \ref{sssec:document_collection} and Figure \ref{fig:source-dist}. Since the  our dataset (\dataset) contains material collected from various web-based sources in the public domain, the copyright and privacy guidelines applied are as specified by these corresponding sources, a few of them as follows: 
\begin{itemize}[leftmargin=*]
    
\item Wikipedia: Text of Creative Commons Attribution-ShareAlike 3.0.\footnote{\footnotesize\url{https://en.wikipedia.org/wiki/Wikipedia:Text_of_Creative_Commons_Attribution-ShareAlike_3.0_Unported_License}}
\item Quora: License and Permission to Use Your Content, Section 3(c).\footnote{\footnotesize\url{https://www.quora.com/about/tos}}
\item Reddit Privacy Policy: Personal information usage and protection.\footnote{\footnotesize\url{https://www.reddit.com/policies/privacy-policy}}
\item Reddit Content Policy.\footnote{\footnotesize\url{https://www.redditinc.com/policies/content-policy}}
\end{itemize}

Future adaptations or continuation of this work would be required to adhere to the policies prescribed herein.

\paragraph{Annotation.}

The annotation was conducted by NLP researchers or linguists in India, who were fairly treated and duly compensated. We conducted several discussion sessions to ensure that all annotators understood the annotation requirements for \expmeme.

\paragraph{Biases.}
Any biases found in the dataset are unintentional, and we do not intend to cause harm to any group or individual. We acknowledge that memes can be subjective, and thus it is inevitable that there would be biases in our gold-labeled data or the label distribution. This is addressed by working on a dataset created using a diverse set of topics and following a well-defined annotation scheme, which explicitly characterizes meme-evidence association. 

\paragraph{Misuse Potential.}
The possibility of being able to deduce relevant contextual, fact-oriented evidence, might facilitate miscreants to modulate the expression of harm against a social entity, and convey the intended message within a meme in an implicit manner. This could be aimed at fooling the regulatory moderators, who could potentially be utilizing a solution like the one proposed to contextualize memes, as such intelligently designed memes might not derive suitable contextual evidence that easily. As a consequence, the miscreants could end-up successfully hindering the overall moderation process. Additionally, our dataset can be potentially used for ill-intended purposes, such as biased targeting of individuals/communities/organizations, etc., that may or may not be related to demographics and other information within the text. Intervention via human moderation would be required to ensure this does not occur.


\paragraph{Intended Use.}

We  curated \dataset\ solely for research purposes, in-line with the associated usage policies prescribed by various sources/platforms. This applies in its entirety to its further usage as well. 
We will distribute the dataset for research purposes only, without a license for commercial use. We believe that it represents a valuable resource when used appropriately.

\paragraph{Environmental Impact.}

Finally, large-scale models require a lot of computations, which contribute to global warming \cite{strubell2019energy}. However, in our case, we do not train such models from scratch; instead, we fine-tune them on a relatively small dataset.
\section*{Acknowledgments}
The work was supported by Wipro research grant.

\bibliography{custom}

\clearpage
\appendix


\section{Implementation Details and Hyperparameter values}
\label{sec:app:imphyper}
We train all the models using Pytorch on an NVIDIA Tesla V100 GPU with 32 GB dedicated memory, CUDA-11.2, and cuDNN-8.1.1 installed. For the unimodal models, we import all the pre-trained weights from the \texttt{\href{https://pytorch.org/vision/stable/models.html}{torchvision.models}} subpackage of the PyTorch framework. We randomly initialize the remaining weights using a zero-mean Gaussian distribution with a standard deviation of 0.02.

We primarily perform manual fine-tuning, over five independent runs, towards establishing an optimal configuration of the hyper-parameters involved. Finally, we train all models we experiment with using the Adam optimizer and a binary cross entropy loss as the objective function.

\begin{table}[t!]
\begin{adjustbox}{width=\columnwidth,center}
\begin{tabular}{c|c|c|c|c|c}
\hline
\textbf{Modality} & \textbf{Model}              & \textbf{BS} & \textbf{EP} & \textbf{\# Param (M)} & \textbf{Runtime (s)} \\ \hline
\multirow{2}{*}{UM} & Bert                        & \multirow{8}{*}{16}          & \multirow{8}{*}{20}          & 110              & 0.66            \\ 
& ViT &            &           & 86               & 0.64            \\ \cline{1-2}\cline{5-6}
\multirow{6}{*}{MM} & Early Fusion                &            &           & 196              & 0.62            \\ 
&CLIP                        &            &           & 152              & 0.73            \\ 
&BAN                         &            &           & 200              & 0.75            \\ 
&VisualBERT                  &            &           & 247              & 0.78            \\ 
&MMBT                        &            &           & 279              & 0.99            \\ \cline{2-6}
&\proposed\                   &            &           & 303              & 0.72            \\ \hline
\end{tabular}
\end{adjustbox}
\caption{\label{tab:hyper-table}Hyper-parameters and per-batch \textit{inference} runtime for each model.}
\end{table}

\section{Additional details about \dataset}
\label{sec:app:sec:mccextra}
\subsection{Meme Collection}
\label{subsec:MemeCollectApp}
We use carefully constructed search queries for every category to obtain relevant memes from the Google Images search engine\footnote{\scriptsize\url{https://images.google.com/}}. Towards searching variants for the topics related \textit{Joe Biden}, some search queries used were ``Joe Biden Political Memes'', ``Joe Biden Sexual Allegation Memes'', ``Joe Biden Gaffe Memes'', ``Joe Biden Ukraine Memes'', among others; for memes related to \textit{Hillary Clinton}, we had ``Hillary Clinton Email Memes'', ``Hillary Clinton Bill Clinton Memes'', ``Hillary Clinton US Election Memes'', ``Hillary Clinton President Memes'', etc. 
For crawling and downloading these images, we use Selenium\footnote{{\scriptsize\url{https://github.com/SeleniumHQ/selenium}}}, a Python framework for web browser automation.

Additionally, for certain categories, we also crawl memes off Reddit. Specifically, We focus on \texttt{r/CoronavirusMemes, r/PoliticalHumor, r/PresidentialRace} subreddits. Instead of using the Python Reddit API Wrapper (PRAW), we use the Pushshift API\footnote{{\scriptsize\url{https://github.com/pushshift/api}}}, which has no limit on the number of memes crawled. We crawl all memes for coronavirus from 1st November 2019 to 9th March 2021. For Biden, Trump, etc., we crawl memes from the other two subreddits and use a set of search queries, a subset of the overall queries we utilized.
\begin{figure}[t!]
    \centering
    \includegraphics[width=\columnwidth]{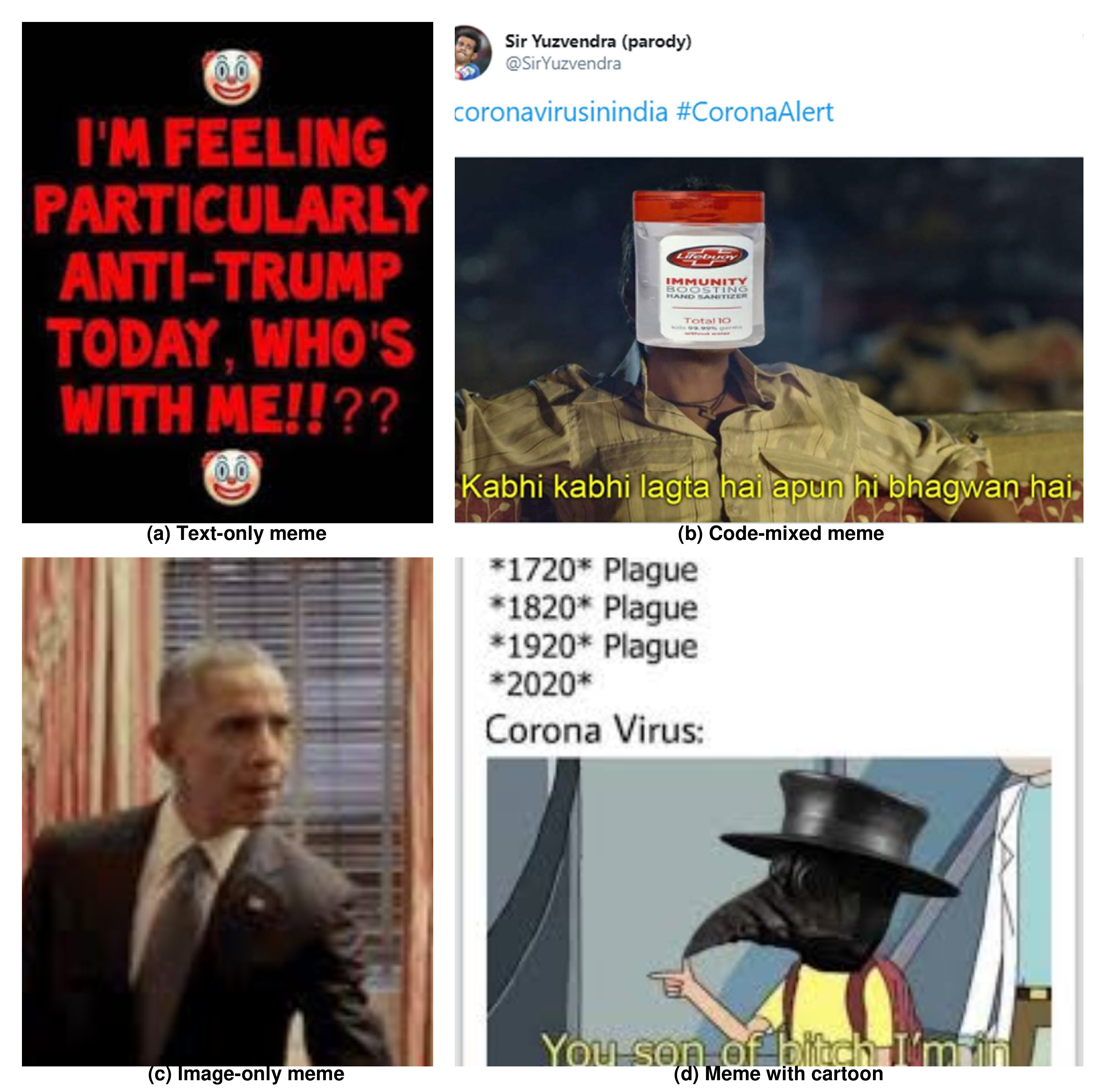}
    \caption{Examples of discarded meme types: (a) Text-only, (b) Code-mixed, (c) Image-only and (d) Cartoon.}
    \label{fig:invalid_memes}
\end{figure}
After scraping all possible memes, we perform de-duplication using dupeGuru\footnote{{\scriptsize\url{https://github.com/arsenetar/dupeguru}}}, a cross-platform GUI tool to find duplicate files in a specified directory. This eliminates intra- and inter-category overlaps. We then remove any meme which is either unimodal, i.e., memes having only images (c.f. Fig. \ref{fig:invalid_memes} (c)), or text-only blocks (c.f. Fig. \ref{fig:invalid_memes} (a)). Additionally, to ensure further tractability of our setup, we manually filter out code-mixed (c.f. Fig. \ref{fig:invalid_memes} (b)) and code-switched memes and memes in languages other than English. Annotating multilingual memes can be a natural extension of our work. We further segregate memes that have cartoons/animations (c.f. Fig. \ref{fig:invalid_memes} (d)). We also filter out memes with poor image quality, low resolution, etc.
\begin{figure*}[t!]
     \centering
     \begin{subfigure}[b]{0.24\textwidth}
         \centering
         \includegraphics[width=\textwidth]{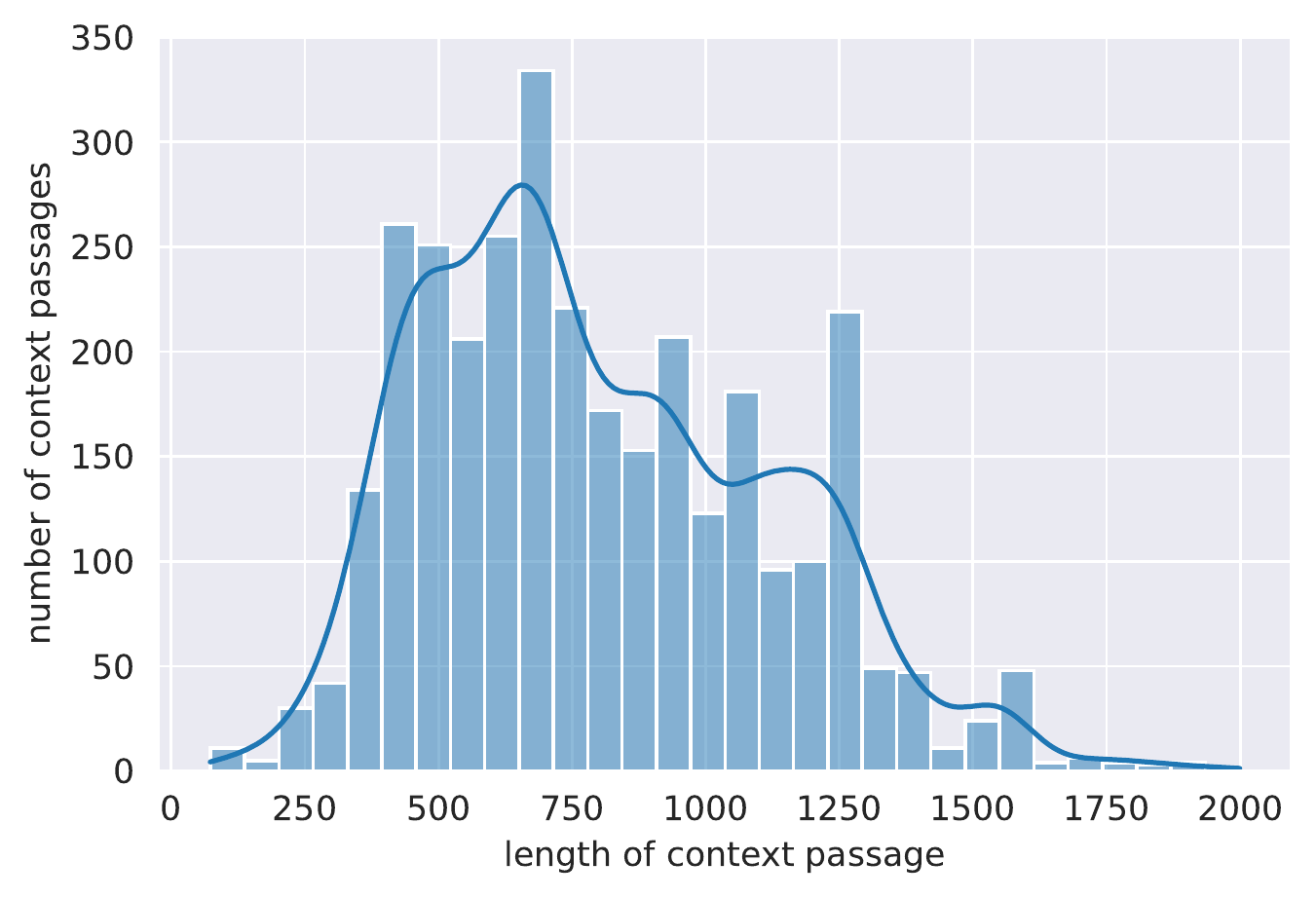}
         \caption{Context length dist.}
         \label{fig:datad}
     \end{subfigure}
     \hfill
     \begin{subfigure}[b]{0.24\textwidth}
         \centering
         \includegraphics[width=\textwidth]{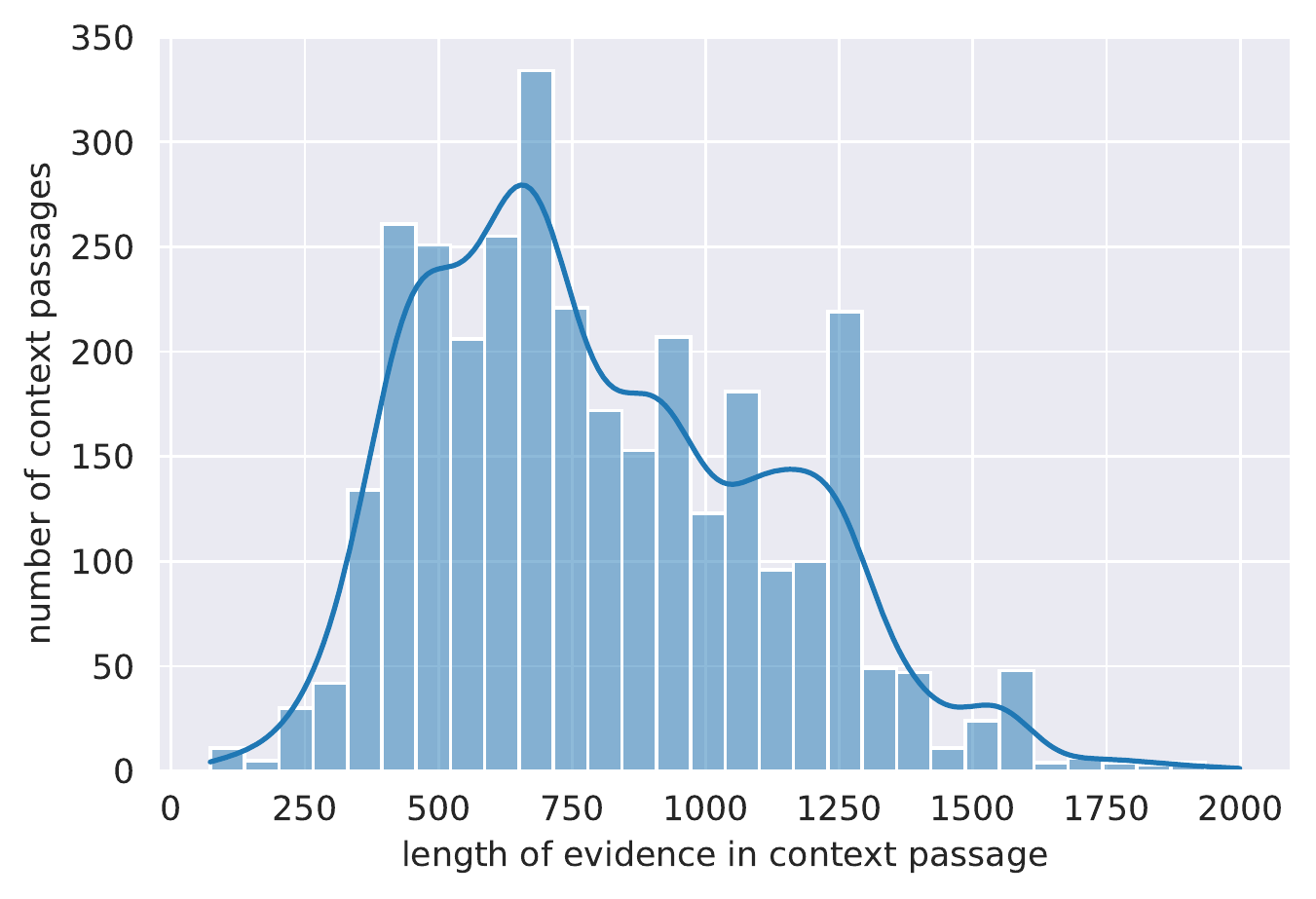}
         \caption{Evidence length dist.}
         \label{fig:datae}
     \end{subfigure}
     \hfill
     \begin{subfigure}[b]{0.24\textwidth}
         \centering
         \includegraphics[width=\textwidth]{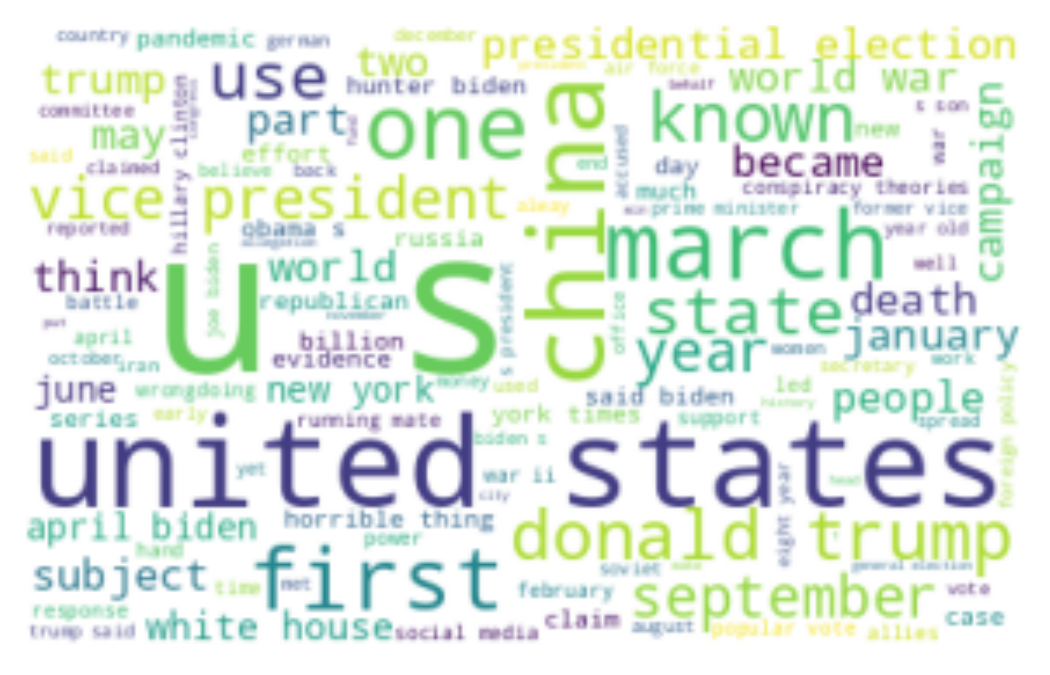}
         \caption{Word Cloud}
         \label{fig:datac}
     \end{subfigure}
     \hfill
     \begin{subfigure}[b]{0.24\textwidth}
         \centering
         \includegraphics[width=\textwidth]{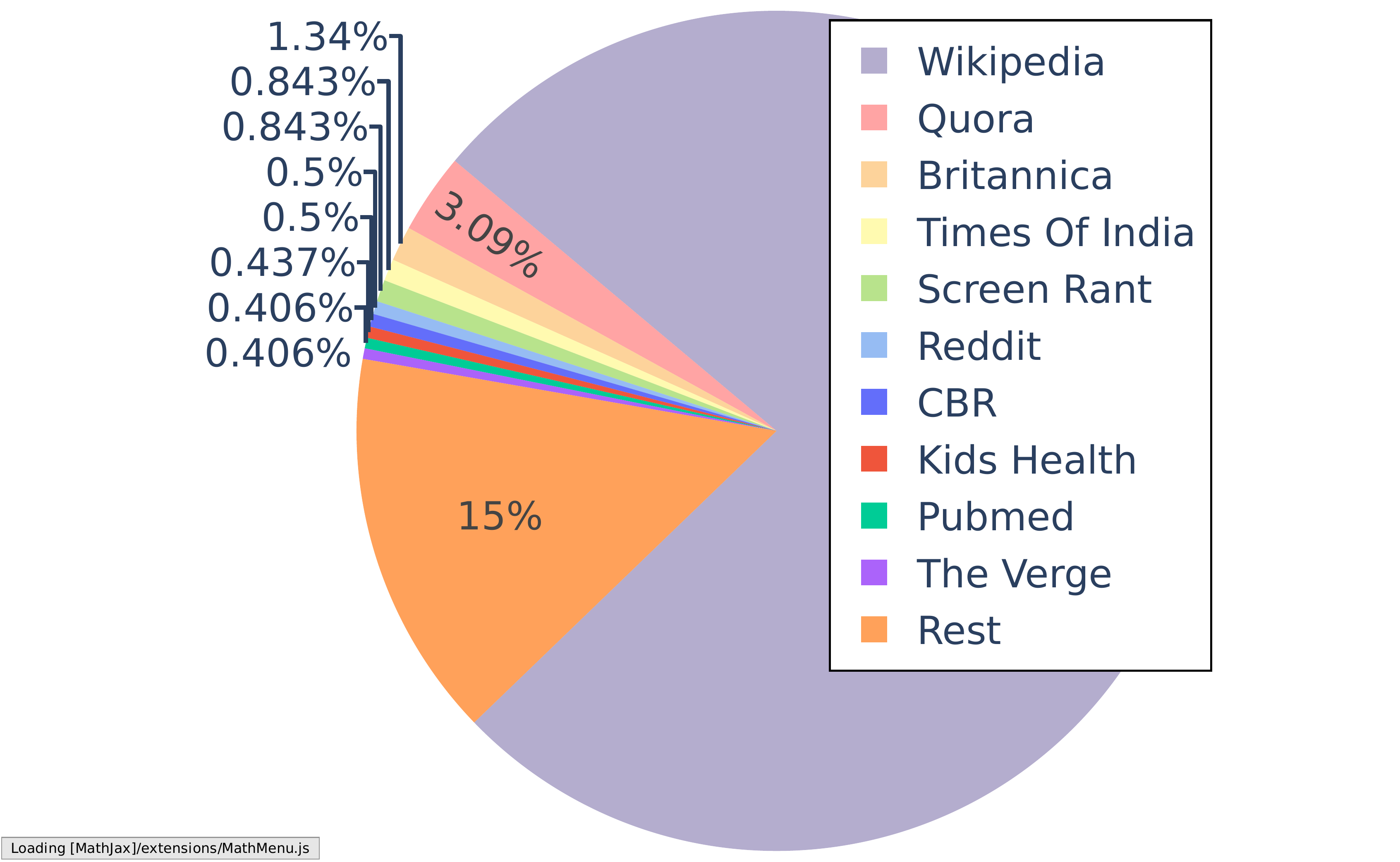}
         \caption{Context source dist.}
         \label{fig:source-dist}
     \end{subfigure}
        \caption{Distribution of attributes in \dataset. The total number of sentences in context passage range between 2 and 16 and the number of evidence sentences in context range between 1 and 10. The most common source of context is Wikipedia.}
        \label{fig:mcc-dist}
\end{figure*}
\subsection{Context Document Curation}
There might be scenarios where: (a) a Wiki document about the topic being reflected in the meme might not exist, or (b) a valid topic-based Wiki page might not contain valid evidence about the information being conveyed within the meme. Since the primary objective of this study is to \textit{investigate and model multimodal contextualization for meme}, we initially mine Wiki documents for topics like ‘politics’ or ‘history,’ for which memes are present online in abundance, thereby leveraging diversity and comprehensiveness facilitated by both the availability of memes and the exhaustive information on a corresponding valid Wiki page. In order to induce generalizability across the topics, types of memes, and context sources, we consider various topics (c.f. Appendix \ref{subsec:MemeCollectApp}) and associated memes and mine the relevant (standard) online information sources (c.f. Fig. \ref{fig:source-dist} towards curating the corresponding context document by performing a Google search for the scenarios where a valid meme-Wiki combination did not hold.

\begin{figure}[t!]
    \centering
    \fbox{\includegraphics[width=0.9\columnwidth]{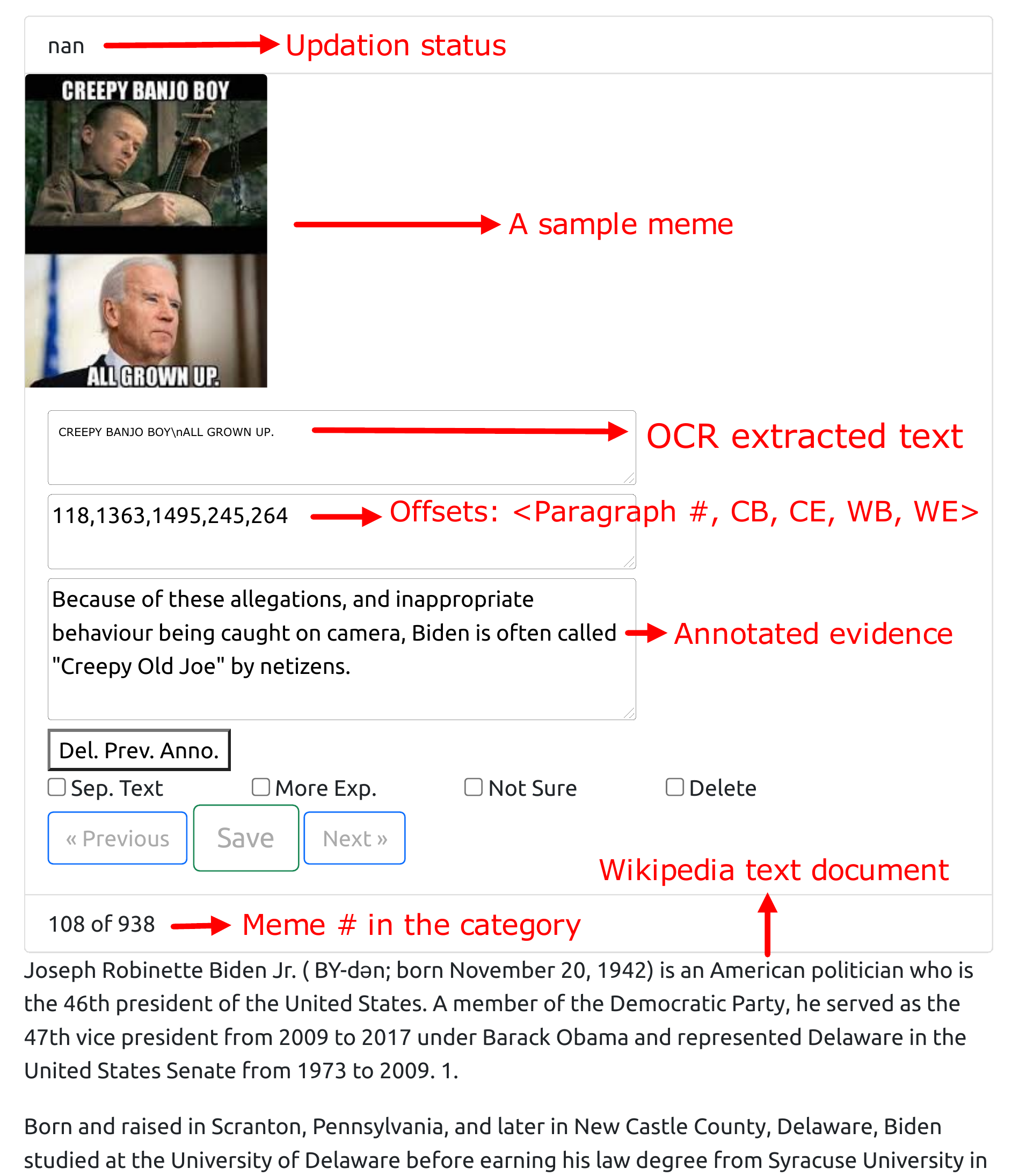}}
    \caption{A Screenshot of the Annotation Tool. Abbr. details for various offsets captured: CB: Character beginning, CE: Character end, WB: Word beginning, WE: Word end.}
    \label{fig:annotation_tool}
\end{figure}

\subsection{Annotation Process}
Two annotators annotated the dataset. One of the annotators was male, while the other was female, and their ages ranged from 24 to 35. Moreover, both of them were professional lexicographers, researchers and social media savvy. Before starting the annotation process, they were briefed on the task using detailed guidelines.

\begin{figure*}[t!]
         \centering
         \includegraphics[width=0.7\textwidth]{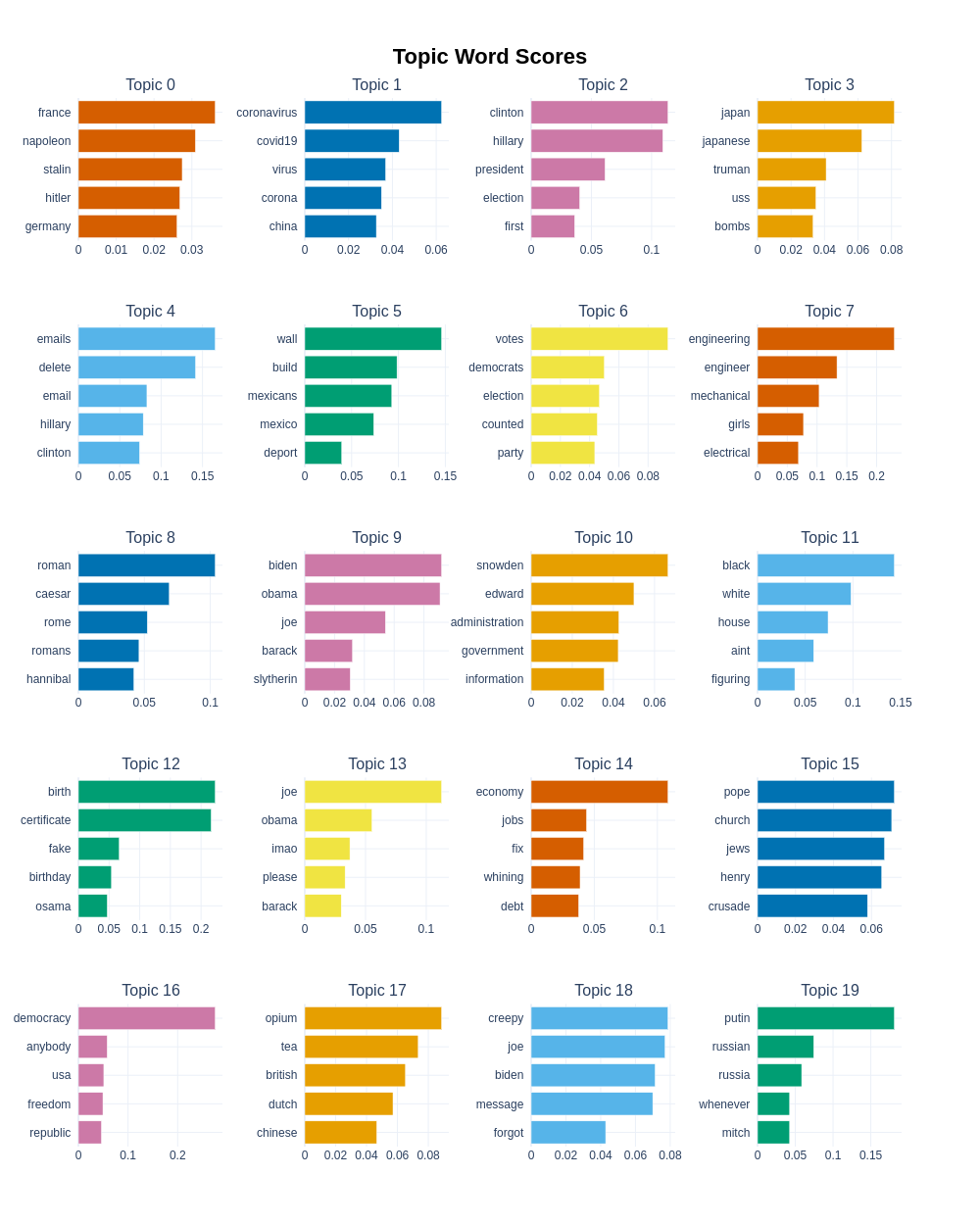}
         \caption{Top-20 prominent topics representing themes of the memetic content in \dataset}
         \label{fig:topic-dist}
\end{figure*}
For performing annotations, we build an annotation platform using JQuery\footnote{{\scriptsize \url{https://github.com/jquery/jquery}}} and Flask\footnote{{\scriptsize \url{https://github.com/pallets/flask}}}. 
A screenshot of the platform is given in Fig. \ref{fig:annotation_tool}. The status of the annotation is displayed at the top. It shows a ``nan'' since the image has not been saved yet; after saving, the status is updated to ``updated''. Below the status, the meme is displayed. There are three text boxes: the first interactive text box is for the OCR text (the annotators can correct and edit the text returned by the OCR pipeline). The other two text boxes are for the offsets and the selected text. The text document in which the explanations are present is at the bottom of the page. When selecting a relevant excerpt from the document, the offsets and selected text are automatically captured and supplemented to the text boxes mentioned above. The format of the offsets, as depicted in Fig. \ref{fig:annotation_tool} is \texttt{<Paragraph Number, Beginning Character Offset, Ending Character Offset, First Word Offset, Last Word Offset>}.

    

\subsection{Analysis and description of \dataset}
It can be observed from Fig. \ref{fig:source-dist} that the highest proportion is from Wikipedia-based sources, followed by smaller proportions for the alternatives explored like Quora, Britannica, Times of India, etc. 
Additionally, the word cloud depicted in Fig. \ref{fig:datac} suggests that most memes are about \textit{prominent US politicians, history, and elections}. Also, context length distribution, as depicted in Fig. \ref{fig:datad}, suggests an \textit{almost} normally distributed context length (in chars), with very few contexts having lengths lesser than $\approx$ 100 and more than $\approx$ 800 chars. Whereas, Fig. \ref{fig:datae} depicts evidence length distribution, according to which most pieces of evidence contain fewer than 400 characters. This corroborates the brevity of the annotated pieces of evidence from diverse contexts. 
\begin{table*}[t!]
    \centering
    \setlength\tabcolsep{1.5pt} 
    \renewcommand{\arraystretch}{0.7} 
    \begin{tabular}{cm{10cm}}
    \hline
    
    \multirow{4}{*}{
    
    \includegraphics[
    width=4.6cm]{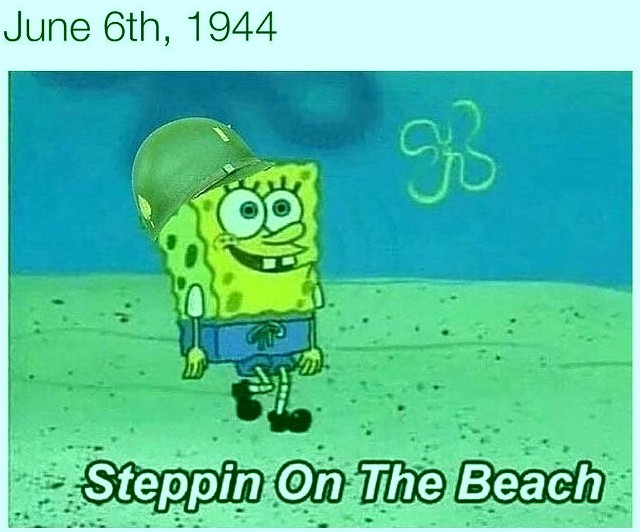}} & \multicolumn{1}{c}{\small \textbf{KYM (\url{knowyourmeme.com})}}\\
    & \tiny \textbf{SpongeBob SquarePants - Steppin On the Beach} (\underline{Image Details})
    \tiny \begin{itemize}[leftmargin=*]
        \item[\ding{112}] 273 views
        \item[\ding{112}] Uploaded 3 years ago
        \item[\ding{112}] Origin Entry: SpongeBob SquarePants (\ding{43} SpongeBob SquarePants is a long-running American television series created by Stephen Hillenburg, airing on Nickelodeon. Learn more on KYM.)
        \item[\ding{112}] Source: \href{https://www.reddit.com/r/dankmemes/comments/bxg1b4/never_forget/}{Reddit}
        \item[\ding{112}] Tags: spongebob, squarepants, dday, history, normandy, landings, neverforget, steppin on the beach, world war ii, france, today in history, nickelodeon, cartoon, reddit, dankmemes
        \item[\ding{112}] About the Uploader: \colorbox{black}{username}, Sr. Researcher \& Scrapbooker \& Media Chauffeur
    \end{itemize} \\ \cdashline{2-2}
     & \multicolumn{1}{c}{\small \textbf{\proposed}}\\
     & \tiny \textcolor{blue}{\textbf{\hl{The Normandy landings were the landing operations and associated airborne operations on Tuesday, 6 June 1944 of the Allied invasion of Normandy in Operation Overlord during World War II}}}. Often referred to as D-Day, it was the largest seaborne invasion in history. The operation began the liberation of France and laid the foundations of the Allied victory on the Western Front. \\ \hline
    \end{tabular}
    \caption{Comparison of the contextual insights obtained from KYM (\url{knowyourmeme.com}, \textit{top}) and the one generated by \proposed\ (\textit{bottom}) for a sample meme. Text blacked-out (\colorbox{black}{X}) is for obscuring the user's identity; Emboldened sentences in blue indicate \textcolor{blue}{\textbf{ground-truth evidences}} and the highlighted sentences indicate \hl{model prediction}.}
    \label{tab:kym_vs_model}
\end{table*}
\subsection{Thematic Analysis from Meme Text}
We perform thematic analysis of the memetic content, using just the text embedded within memes. We took the OCR extracted meme's text and project top-20 topics using \textit{BERTopic} \cite{grootendorst2022bertopic}, a neural topic modeling approach with a class-based TF-IDF procedure.

We depict 0-based topic indexes and thematic keywords as 0--History, 1--Covid-19, 2--Politics, 3--War with Japan, etc., in Fig. \ref{fig:topic-dist}. These topics are collectively referenced and described via the most likely keywords appearing for that particular topic. This depiction also highlights how generalizable our proposed approach is in optimally detecting accurate evidence from various topics within a given related context. Besides different high-level topics, \dataset\ also captures the diversity of the sub-topics. Although, except for a few topics like Topics: 15 and 18, reasonably diverse memes can be found in \dataset. 
\section{Comparing contexts from KYM and \proposed}
\label{sec:appkymmime}

Here, we compare the insights available on \url{knowyourmeme.com} (also referred to by KYM) and the ones generated by our proposed modeling framework \proposed,\ about a particular meme. For comparison, we consider a sample meme (c.f. Table \ref{tab:kym_vs_model}) from our test set, which also happens to be available on KYM\footnote{\scriptsize\url{https://knowyourmeme.com/photos/1500530-spongebob-squarepants}}. This meme is about a soldier (portrayed via character \textit{SpongeBob}) stepping onto the beach on June 6th, 1944, which is an implicit reference to the D-Day landings during World War II. We present our comparative analysis in the following subsections.  

\subsection{\proposed}
Since Wikipedia articles are supposed to document in-depth factual details related to events, people, places, etc., one can expect the information obtained to be exhaustive, which is what \proposed\ aims to leverage. \proposed\ achieves this by establishing a cross-modal evidence-level association between memes and a supplementary context document. While there are different levels of details (with varying relatedness) present within Wikipedia documents, there are one or more sentences that \textit{distinctly complement} the meme's intended message. 

In this case, the excerpts emboldened and highlighted contribute to building the meme's rationale, as depicted in Table \ref{tab:kym_vs_model}. The key advantages to using an approach like \proposed\ can be enlisted as follows:
\begin{itemize}[leftmargin=*]
    \item Information derived can facilitate comprehensive assimilation of the meme's intended message.
    \item \proposed\ does not rely on manually archived details and meta-data. Instead, it presumes the availability of a \textit{related} context, which can be easily mined from the web.
    \item Finally, \proposed\ can optimally detect accurate contextual evidence about a meme without presenting information that might not be useful.
\end{itemize}
Although \proposed\ in its current stage has limitations, it would require active fine-tuning and optimization towards regulating its cross-modal associativity, towards modeling memetic contextualization.

\subsection{KYM}
On the other hand, as can be observed from Table \ref{tab:kym_vs_model}, KYM divulges the details like (a) total views, (b) time of upload, (c) origin details, (d) source, (e) relevant tags and (e) up-loader details. Most of this information could be considered as meta-data, w.r.t. the meme (template). Such multimedia information captures the details related to its digital archival. The following factors characterize such information:
\begin{itemize}[leftmargin=*]
    \item The \textit{origin} information about a meme is likely to be one of the most critical information, as it presents details regarding the inception of a particular meme, which is often imperative to establish the underlying perspective conveyed within a meme.
    \item Although \textit{tags} aggregate a comprehensive set of related entities, it can also include some irrelevant information.
    \item Other available meta-data like \textit{no. of views, date uploaded}, etc., could be beneficial w.r.t. detecting meme's harmfulness or virality over social media, but not as much towards divulging meme's intended message.
\end{itemize}
Information provided by KYM \textit{may or may not} be sufficient to comprehend the actual meme's intended message, as it significantly relies on human intervention towards curating such data and is therefore always bound to be limited. Still, information like the \textit{origin details} and \textit{related tags} can facilitate establishing the mappings across layers of abstraction that memes typically require.


\end{document}